\documentclass[letterpaper]{article} 
\usepackage{aaai2026}  
\usepackage{times}  
\usepackage{helvet}  
\usepackage{courier}  
\usepackage[hyphens]{url}  
\usepackage{graphicx} 
\usepackage{natbib}  
\usepackage{caption} 
\usepackage{algorithm}
\usepackage{algorithmic}

\usepackage{newfloat}
\usepackage{listings}
\DeclareCaptionStyle{ruled}{labelfont=normalfont,labelsep=colon,strut=off} 
\floatstyle{ruled}
\newfloat{listing}{tb}{lst}{}
\floatname{listing}{Listing}
\usepackage{tabularx}
\usepackage{multirow}
\usepackage{booktabs}
\usepackage{makecell}
\usepackage{amssymb}
\usepackage[most]{tcolorbox}
\usepackage{hhline}
\usepackage{longtable}
\usepackage{enumitem}
\usepackage{xcolor}

\newcolumntype{Z}{>{\centering\let\newline\\\arraybackslash\hspace{0pt}}X}
\newcolumntype{P}[1]{>{\centering\arraybackslash}p{#1}}

\definecolor{resolu_color}{RGB}{26, 158, 0}

\title{StoryBox: Collaborative Multi-Agent Simulation for Hybrid Bottom-Up Long-Form Story Generation Using Large Language Models}
\author{
    Zehao Chen\textsuperscript{\rm 1},
    Rong Pan\textsuperscript{\rm 1}\thanks{Corresponding authors.},
    Haoran Li\textsuperscript{\rm 2}\footnotemark[1]
}
\affiliations{

    \textsuperscript{\rm 1}School of Computer Science and Engineering, Sun Yat-sen University, Guangzhou, China \\
    \textsuperscript{\rm 2}School of Arts, Sun Yat-sen University, Guangzhou, China \\
    chenzh593@mail2.sysu.edu.cn, 
    panr@sysu.edu.cn,
    lihr83@sysu.edu.cn
}

\begin{document}

\maketitle

\begin{abstract}
  Human writers often begin their stories with an overarching mental scene, where they envision the interactions between characters and their environment. Inspired by this creative process, we propose a novel approach to long-form story generation, termed hybrid bottom-up long-form story generation, using multi-agent simulations. In our method, agents interact within a dynamic sandbox environment, where their behaviors and interactions with one another and the environment generate emergent events. These events form the foundation for the story, enabling organic character development and plot progression. Unlike traditional top-down approaches that impose rigid structures, our hybrid bottom-up approach allows for the natural unfolding of events, fostering more spontaneous and engaging storytelling. The system is capable of generating stories exceeding 10,000 words while maintaining coherence and consistency, addressing some of the key challenges faced by current story generation models. We achieve state-of-the-art performance across several metrics. This approach offers a scalable and innovative solution for creating dynamic, immersive long-form stories that evolve organically from agent-driven interactions.
\end{abstract}

%
\begin{links}
    \link{Project}{https://storyboxproject.github.io}
\end{links}

\section{Introduction}

\begin{figure}[t]
  \centering
  \includegraphics[width=\columnwidth]{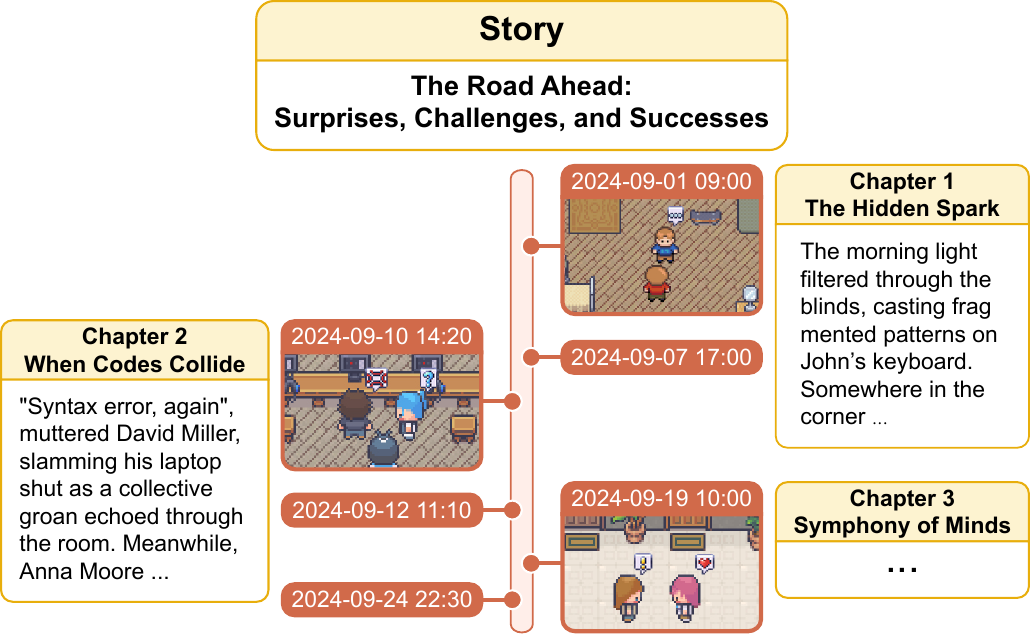}
  \caption{The timeline of the multi-agent sandbox simulation, where agent interactions with each other and their environment trigger emergent events that drive dynamic, hybrid bottom-up story generation.}
  \label{fig:timeline}
\end{figure}

The advent of large language models (LLMs) brings significant advancements to various fields, including multi-agent simulations \cite{li-etal-2024-econagent, qian-etal-2024-experiential, chen-etal-2024-llmarena}. These simulations offer a powerful tool for modeling complex interactions between virtual agents, providing a dynamic and context-rich environment for story generation. When humans write stories, they typically have an overarching mental picture of the story world. In our approach, multi-agent simulations are used to create this overarching scene, where agents (i.e., the characters in the story) interact based on predefined behaviors, triggering emergent events that form the foundation of the story. This allows for the automatic generation of diverse and engaging scenarios, which are crucial for building long-form stories.

As shown in Figure~\ref{fig:timeline}, the multi-agent sandbox simulation unfolds over time, with a series of events occurring within the sandbox. These events arise from interactions between agents, as well as between agents and their environment. In this process, agents act based on their predefined attributes, responding to and influencing the world around them. The interactions between agents and their environment give rise to a chain of events that continuously evolve, creating a dynamic storyline. The sandbox thus serves as a virtual space that mirrors the mental scene a human writer might envision when crafting a story. It is within this evolving sandbox that the foundations of the story are formed, from which a complete story can be generated based on the events that unfold.

In this setup, the sandbox serves a critical function: it is akin to the mental scene which a writer imagines, but with the difference that the interactions and outcomes are automatically generated through agent-based simulation rather than purely through human imagination. Each interaction, whether between characters or between characters and their surroundings, adds new layers to the story. By allowing agents to operate in this evolving and flexible sandbox, we ensure that the story's events are not preordained but emerge organically from the agents' behavior and environmental dynamics.

Traditional story generation often follows a top-down method \cite{zhou2023recurrentgpt, wang-etal-2023-improving-pacing, fan-etal-2019-strategies, goldfarb-tarrant-etal-2020-content, yang-etal-2022-re3}, where a story's structure is outlined first and then expanded. Although this framework provides a guide, it can limit natural character development and plot progression, resulting in stories that feel forced or predictable. In contrast, our hybrid bottom-up approach starts with agent-driven simulations that generate emergent events. As agents interact, these events unfold naturally, enabling more organic character growth and plot evolution, ultimately producing stories with greater depth, coherence, and spontaneity.

Our approach excels at generating longer, more detailed stories that stay consistent and cohesive. Using multi-agent simulations, it creates diverse, meaningful events that enrich plot and characters. Our main contributions are:

\begin{itemize}
  \item A multi-agent simulation framework that generates dynamic and contextually rich story events through interactive agent behaviors.
  
  \item A hybrid bottom-up process that combines emergent interactions with guided storytelling to enable natural character development and plot progression.
  
  \item The ability to produce coherent and consistent long-form stories exceeding 10,000 words, effectively overcoming limitations of current story generation models.
\end{itemize}

\section{Related Work}

\subsection{LLM-Based Multi-Agent Simulation}

LLM-based multi-agent simulations have gained attention for their advanced language processing and decision-making, enabling nuanced agent interactions \cite{jinxin2023cgmi, liu2023training, feng2023role}. These systems have been explored in various domains \cite{wang2024survey, williams2023epidemic}, particularly social simulation, which is most relevant to our research. Social simulation with LLM-based agents provides an effective way to model complex societal dynamics that are otherwise difficult or costly to study \cite{li2023you, li2023quantifying}. For example, S$^3$ \cite{gao2023s} investigates the spread of information, emotions, and attitudes in social networks, while Generative Agents \cite{park2023generative} and AgentSims \cite{lin2023agentsims} model daily human interactions in virtual towns. Social Simulacra \cite{park2022social} focuses on simulating online community regulation, and SocialAI School \cite{kovavc2023socialai} uses LLMs to simulate child development. These studies highlight the growing potential of LLM-based agents to offer valuable insights into societal behavior, community regulation, and social development, which provides a foundation for hybrid bottom-up long-form story generation.

\subsection{Story Generation}

Story generation has advanced with the rise of LLMs, which produce fluent stories but often struggle with coherence and consistency \cite{huot2024agents}. To address these challenges, frameworks such as outlining followed by expansion into full stories have been proposed \cite{zhou2023recurrentgpt, wang-etal-2023-improving-pacing}. However, long-form story generation remains challenging. Recent studies have explored multi-agent systems using LLMs \cite{wang2024storyverse, nasir2024word2world}, such as Agents' Room \cite{huot2024agents}, which focuses on generating stories of 1,000-2,000 words using Planning and Writing Agents coordinated by an Orchestrator, and IBSEN \cite{han-etal-2024-ibsen}, which employs Director-Actor agent collaboration for scriptwriting. While top-down approaches, such as outlining, provide structure, they limit natural storylines and character interactions. In contrast, our hybrid bottom-up approach generates dynamic stories from simulations, producing richer and more coherent stories of over 10,000 words, with clear advantages over traditional methods.

\section{Methodology}

\begin{figure*}[t]
  \centering
  \includegraphics[width=\textwidth]{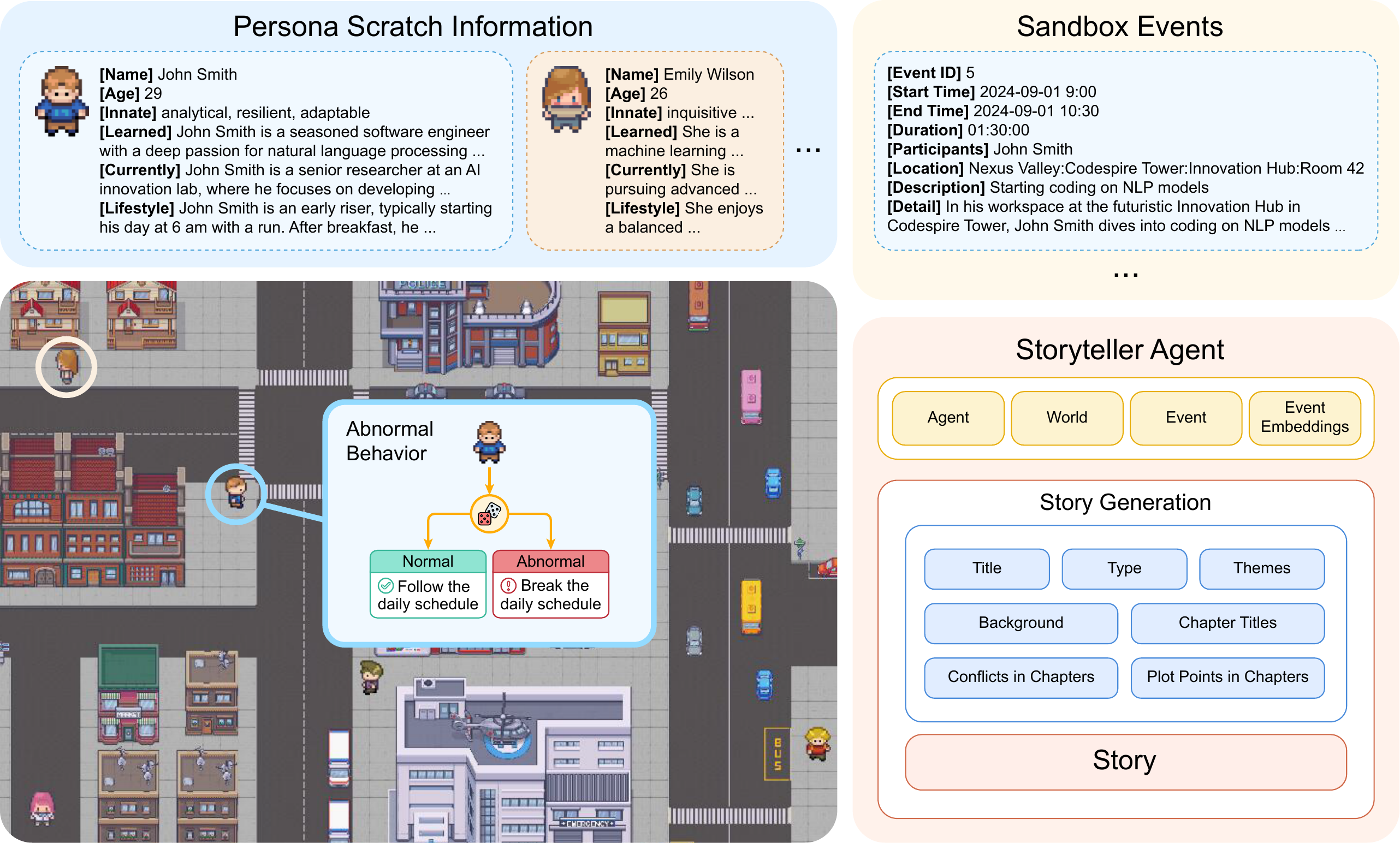}
  \caption{Overview of the system framework for long-form story generation, including the Persona Scratch Information for defining character settings, the sandbox where agent interactions generate events, and the Storyteller Agent that uses these events to craft a complete story.}
  \label{fig:framework}
\end{figure*}

This section introduces our long-form story generation system (Figure~\ref{fig:framework}). It has two parts: a sandbox simulating multi-agent interactions to create events, and a Storyteller Agent that turns these events into coherent, engaging stories.

\subsection{Multi-Agent Simulation}

The multi-agent simulation serves as the foundation for generating long-form stories. Both the process of long-form story generation and standalone multi-agent simulation rely on well-defined character settings. Inspired by the character modeling approach of Generative Agents \cite{park2023generative}, we adopt and modify it to better suit the specific requirements of story generation. This modification ensures that the characters exhibit coherent behavior and contribute meaningfully to the overall story structure.

\subsubsection{Core Attributes}

As illustrated in Figure~\ref{fig:framework} ``Persona Scratch Information'', we first define the initial settings for each character. These initial settings are critical for the agents to engage in realistic interactions and generate compelling events. Each character is defined by a set of core attributes, which include Name, Age, Innate, Learned, Currently, and Lifestyle. These attributes describe both static and dynamic aspects of the character, providing a foundation for their behaviors and decisions. To model the character's daily actions, we introduce an additional attribute, Daily Plan Requirements. This attribute contains specific tasks or routines that the character plans to accomplish, such as conducting research, reading papers, etc. Importantly, this attribute is dynamic: at the start of a new day, a fresh set of daily plans is generated. Based on these daily plan requirements, a detailed schedule is created, with tasks assigned to specific hours of the day. This enables the character to follow a structured routine, while leaving room for flexibility and variation.

\subsubsection{Abnormal Behavior Attribute}

To add depth and excitement to the generated stories, it is crucial to incorporate elements that break from routine. A monotonous, predictable character would lead to dull and uneventful stories. Therefore, we introduce an Abnormal Behavior attribute for each character. This attribute determines whether the character is likely to engage in actions that deviate from their usual behavior, such as abandoning their daily plan to pursue other activities or engaging in conflicts with other characters. The probability of a character exhibiting abnormal behavior is controlled by a hyperparameter, the Abnormal Factor. A higher value for this factor increases the likelihood that the character will break from their routine and introduce more dynamic and unpredictable actions.

\subsubsection{Character Behavior Types}

Character behavior is categorized into three main types: move, chat, and none. The move behavior indicates that the character physically moves to a different location and performs specific actions. The chat behavior represents the character engaging in conversations with other characters. The none behavior means the character does nothing, allowing for moments of reflection, rest, or inaction, which contribute to pacing and tension within the simulation. These behaviors are selected based on the character's current attributes and the context of the simulation.

Together, these attributes define a character's personality, daily routines, and potential for unpredictability, creating a rich and dynamic environment for simulation. The agents interact with each other based on these characteristics, generating events that reflect plausible interpersonal dynamics and responses to environmental changes. This simulation framework serves as the raw material for long-form story generation, with the emergent events providing the foundation for compelling stories.

\subsubsection{Event Recording}

In the sandbox, every action performed by an agent is considered an event, as illustrated in Figure~\ref{fig:framework} under ``Sandbox Events''. Each event is assigned a unique ID within the sandbox, along with a defined start time and end time, allowing us to calculate its duration. Events can involve multiple participants; for example, a conversation event will always have at least two participants. Additionally, each event must be recorded with its location, a brief description of the event (referred to as description), and a more detailed account (referred to as detail) to capture the full context.

The description attribute provides a concise summary of the event, such as ``Starting coding on NLP models'', which briefly explains what the event is about. However, this brief description is often insufficient for generating a rich and engaging story, as it lacks important contextual details. To address this, we introduce the detail attribute, which expands upon the basic description and includes additional information necessary for story generation. The detail attribute incorporates factors such as the environment, the time of day, the status of the characters involved, and the location of the event. Furthermore, it captures character-specific elements, including their actions, emotional states, and possible motivations during the event.

By providing these details, we ensure that the events within the sandbox are not just isolated actions, but are rich, contextually grounded occurrences that can later be woven into the story. This level of granularity allows the Storyteller Agent to craft stories that are nuanced and immersive, as the events are not only linked to character behaviors but are also informed by the surrounding context and dynamics.

Through this multi-agent simulation, we create a dynamic sandbox environment where interactions between characters generate a wide variety of events. These events, rich in context and detail, form the core elements for story generation. The Storyteller Agent then processes these events, transforming them into intricate, character-driven long-form stories that maintain both depth and coherence in the story.

\subsubsection{Environment Modeling}

\begin{figure*}[t]
  \centering
  \includegraphics[width=\textwidth]{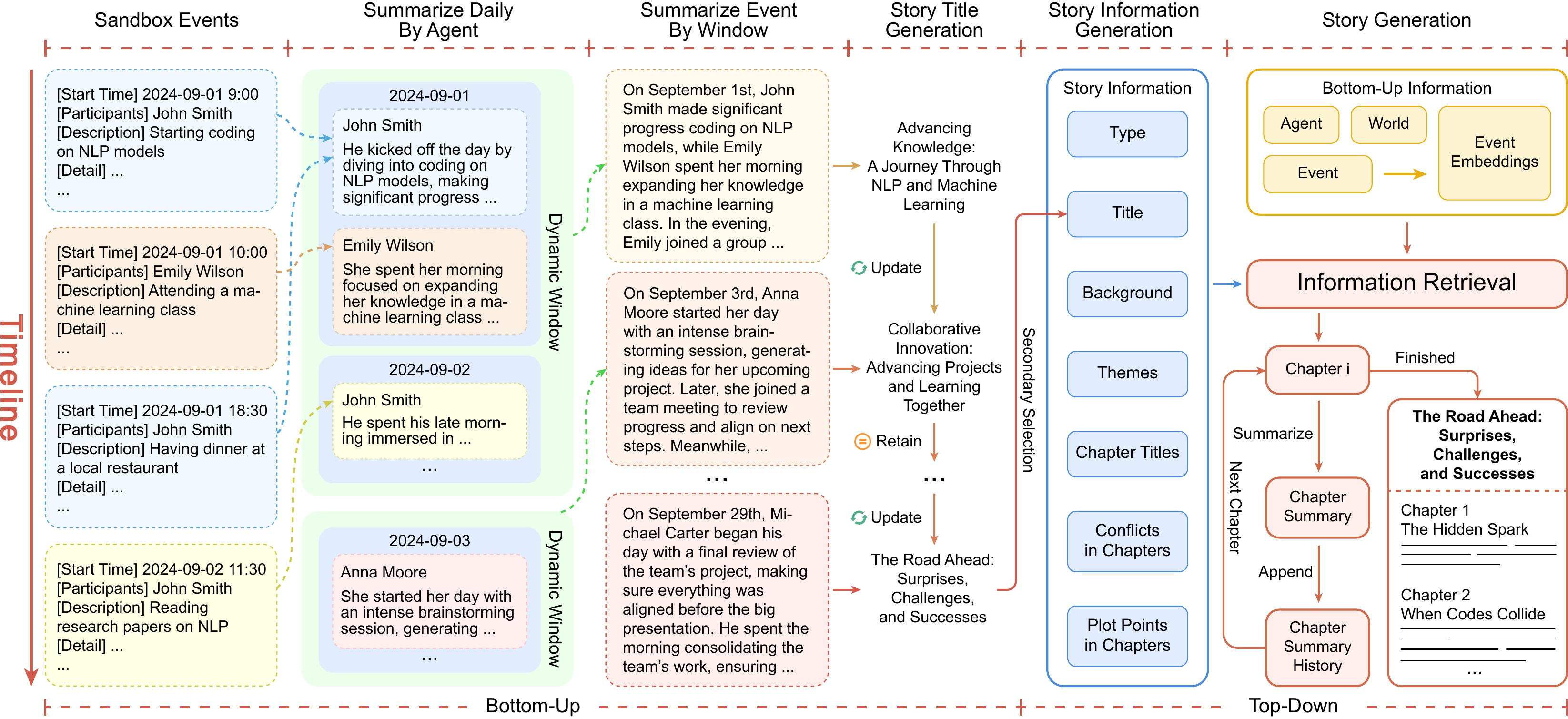}
  \caption{Overview of the Storyteller Agent workflow for generating long-form story using a hybrid bottom-up approach, from sandbox events to iterative story generation.}
  \label{fig:storyteller}
\end{figure*}

The environment plays a critical role in the sandbox. In Generative Agents, the environment is modeled using a tile-based system, where tiles represent the basic environmental units commonly found in RPG games. However, this approach has inherent limitations, particularly the need for precise coordinates, which can make modeling more rigid and less flexible. To overcome these limitations, we adopt a more general tree-like structure for environment modeling, which does not rely on specific coordinates but instead uses relative distances between objects and areas. For a detailed introduction to environment modeling, please refer to the supplementary material.

Unlike the fixed, grid-based environment of Generative Agents, which confines agents to a limited simulation space, our model enables a virtually limitless environment. This means that the sandbox can be as large and complex as needed, without the constraint of pre-defined spatial boundaries. Characters can explore a more expansive and immersive world, interacting with a wider range of objects and areas, making the simulation more dynamic and fluid.

\subsection{Hybrid Bottom-Up Long-Form Story Generation}

Once the events generated through the multi-agent sandbox simulation are available, the next step is to use the Storyteller Agent to generate the long-form story. As illustrated in Figure~\ref{fig:storyteller}, this process follows a structured workflow that incorporates several key stages.

\subsubsection{Event Summarization}

The sandbox events are arranged chronologically, but the sheer number of events can be overwhelming. Directly feeding all these events into a LLM is not feasible due to the limitations of the model's context window. Therefore, summarization is necessary. First, we summarize the events by character, recording what each character did on each day. This summary is also arranged chronologically. Once we have these character-based daily summaries, we introduce a dynamic windowing mechanism that groups events into smaller chunks for summarization. The size of the dynamic window is automatically determined by the LLM, allowing for adaptability to the complexity and density of the events. The summaries within each window are abstract and condensed, reducing the volume of data while maintaining chronological order. This layered and dynamic approach ensures that the event data remains manageable for the LLM while preserving key information.

\subsubsection{Story Information Generation}

In the story generation process, we first generate the story type (e.g., adventure, mystery, romance) rather than beginning with the title. The story type plays a crucial role in shaping the story, as it provides a foundational framework that guides the subsequent development of the plot, characters, and themes. By determining the story type at the outset, we establish a clear direction for the story, ensuring coherence and consistency throughout the story's progression.

After determining the story type, we generate the story title beginning with the initial event summary. The title is iteratively refined with each new event summary, allowing the model to update the title dynamically based on evolving context. This process not only produces a cohesive and relevant title but also acts as a filter, highlighting key elements and capturing the story's essence. The language model decides whether to retain or modify the title depending on the current event summaries.

Once the title is finalized, a secondary selection process identifies the most suitable title. This can involve using a language model for evaluation, manual review, or training a dedicated model. Due to the lack of a specialized dataset, we opt for a fully automated LLM-based approach that ranks titles based on relevance, creativity, and coherence with the event summaries.

Following title generation, other key story information is produced, including background, main themes, chapter titles, conflicts per chapter, and major plot points. These elements are informed by the story type, title, and prior event summaries, ensuring a well-structured and thematically consistent story.

If manual inputs are provided, the Storyteller Agent skips automatic generation of these elements. Additionally, hyperparameters like the number of themes, chapters, conflicts per chapter, and plot points per chapter allow customization of the story's scope and depth.

\subsubsection{Story Generation}

Once the story information is ready, the actual story generation process begins. A key component in this stage is the information retrieval module, which receives two types of input data: story-related information (such as title, themes, and chapter details) and sandbox data (including character information, environment details, and the events themselves). To further enhance event matching for information retrieval, we convert events into event embeddings using an embedding model. This allows events to be retrieved both through keyword-based search and dense vector-based retrieval.

It is important to highlight that the process of information retrieval and its application in story generation is inherently bottom-up in nature. This is because the events from the sandbox, which play a crucial role in the generation process, are drawn upon, distinguishing this approach from traditional top-down story generation methods. Furthermore, this information retrieval process also acts as a dynamic filtering mechanism, automatically selecting meaningful events that align with the story's progression. By continuously refining the event selection based on the evolving story, the system ensures that only the most relevant and engaging content is used to shape the story.

The story generation process is driven by an iterative loop. During the generation of the first chapter, the system retrieves relevant information, such as the story title, themes, and chapter details, to serve as the foundation for crafting the content. It's important to note that a chapter may not be fully generated in a single pass; instead, it undergoes several iterations before it is completed. Once a chapter is finished, it is summarized, and the summary is added to the chapter summary history, which serves as input for generating the next chapter.

Each subsequent chapter's generation includes not only the information retrieved for the current chapter but also the summaries of the previous chapters. This iterative process continues, chapter by chapter, until the entire story is completed. This hybrid bottom-up method allows for the generation of long, coherent stories, where each chapter builds on the events and summaries that have come before, ensuring continuity and story flow.

\section{Experiments}

In this section, we present experiments to validate the effectiveness of StoryBox, including evaluation metrics, baseline comparisons, and result analysis.

\begin{figure*}[ht]
  \centering
  \includegraphics[width=\textwidth]{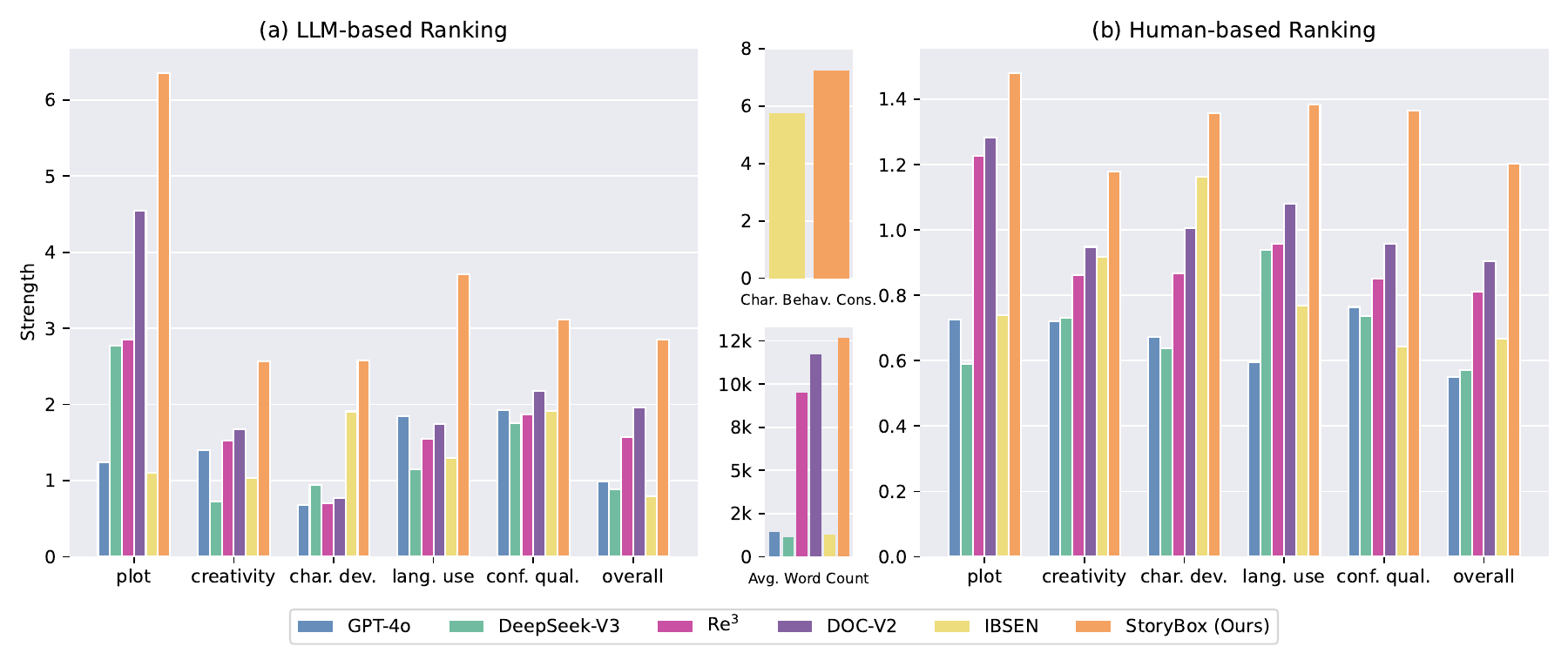}
  \caption{Comparative performance of different methods across multi evaluation dimensions: Plot, Creativity, Character Development, Language Use, Conflict Quality, and Overall. Subfigure (a) presents rankings based on LLM-based evaluation, while subfigure (b) shows rankings from human evaluation. We also include the sandbox-specific metric Character Behavior Consistency, along with the Average Word Count for each method.}
  \label{fig:performance}
\end{figure*}

\subsection{Experimental Settings}

\subsubsection{Dataset}

Following the setup described in DOC \cite{yang-etal-2023-doc}, we use premises, settings, and characters as inputs. However, the DOC dataset contains relatively homogeneous story types, so we construct a new dataset consisting of 20 story-related settings without reference answers. Our dataset features a broader variety of story types, including genres such as science fiction. For detailed information about our dataset, please refer to the supplementary material.

\subsubsection{Evaluation Metrics}

Evaluating generated stories is inherently challenging due to the subjective, multi-dimensional, and context-dependent nature of narrative quality \cite{guan-huang-2020-union, chhun-etal-2022-human, yang2024makes}. Inspired by recent advances such as Agents' Room \cite{huot2024agents}, we adopt a hybrid evaluation strategy that combines both human and automated methods to provide a comprehensive assessment of story quality. Building upon this foundation, we expand the evaluation criteria to better suit the story generation setting.

\paragraph{(1) Human Evaluation}

We conduct structured pairwise comparisons across six criteria: Plot, Creativity, Character Development, Language Use, Conflict Quality, and Overall. These dimensions cover key aspects of long-form storytelling, including narrative coherence, originality, emotional depth, and stylistic quality. Annotators with backgrounds in literature, storytelling, and related fields compare story pairs from the same setting and provide ratings for each dimension as well as an overall preference. To reduce fatigue and bias, each annotator evaluates a limited number of pairs per session. We apply a Latin Square design to balance the assignment of story settings and randomize story order. For each setting, we collect $N \times (N - 1) / 2$ comparisons, where $N$ is the number of methods.

\paragraph{(2) Automatic Evaluation}

We also use a large language model to perform evaluations based on the same criteria as human judgments. The model is guided by structured prompts aligned with human instructions and compares story pairs across key dimensions. This approach offers a cost-effective approximation of human evaluation and allows us to assess its alignment with human ratings. In addition, we include a sandbox-specific metric, \textbf{Character Behavior Consistency}, which evaluates how well characters act in line with their defined personas. We also report \textbf{Average Word Count} to reflect story length.

For full metric definitions, evaluation procedures, and interface details, please refer to the supplementary material.

\subsubsection{Baselines}

We select several representative and high-performing methods as our baselines. We categorize the baseline methods into three types: (1) \textbf{Vanilla LLMs}: methods that generate long-form stories directly using large language models, such as GPT-4o \cite{hurst2024gpt} and DeepSeek-V3 \cite{liu2024deepseek}; (2) \textbf{Structured Frameworks}: methods that utilize structured frameworks for long-form story generation based on LLMs, such as Re$^3$ \cite{yang-etal-2022-re3} and DOC-V2 \cite{yang-etal-2023-doc}; and (3) \textbf{Multi-Agent Simulations}: methods that generate through multi-agent simulations, such as IBSEN \cite{han-etal-2024-ibsen}.

\subsubsection{Implementation Details}

The process of implementing the system, including the story-related settings, sandbox initialization, and the setup of characters and environment, is described in the supplementary material. The specific prompts used in this paper, along with additional implementation details, are also provided therein.

\subsection{Automatic Evaluation}

We conduct an automatic evaluation of several methods, with results shown in Figure~\ref{fig:performance}(a). StoryBox consistently achieves the best performance across all major metrics.

For the \textbf{Plot} metric, it maintains strong coherence, while IBSEN performs less well, likely due to its lack of narrative-specific design. Most models show similar results in \textbf{Creativity}, except DeepSeek-V3, which performs noticeably worse, suggesting limited stylistic diversity. In \textbf{Character Development}, StoryBox ranks highest, and IBSEN follows, indicating that sandbox-based simulations are effective in modeling character dynamics. StoryBox also leads in \textbf{Language Use}, thanks to the Storyteller Agent's integration of environmental and contextual cues, which enriches the narrative with vivid descriptions. For \textbf{Conflict Quality}, it performs best, reflecting its ability to construct structured and engaging tension. The \textbf{Overall} score confirms StoryBox's strong and balanced performance across dimensions. In the sandbox-specific \textbf{Character Behavior Consistency} metric, StoryBox slightly outperforms IBSEN, suggesting that simulation contributes to coherent agent behavior. Finally, in the \textbf{Average Word Count} metric, which reflects long-form generation ability, StoryBox produces stories averaging around 12,000 words. Other long-form methods typically generate about 10,000 words, while models without such adaptation produce much shorter texts, averaging only around 1,000 words, revealing a clear limitation in generating extended story.

Additional results, including case studies, are provided in the supplementary material.

\subsection{Human Evaluation}

Given the inherently subjective nature of story generation, human evaluation is essential for a comprehensive assessment. To capture diverse reader perspectives, we recruited 78 participants from a variety of academic backgrounds, including literature, computer science, and other disciplines. Each participant was asked to compare pairs of stories generated from the same setting, based on the same set of dimensions used in the automatic evaluation.

The human evaluation results are presented in Figure~\ref{fig:performance}(b). Overall, StoryBox consistently achieves the highest scores across all dimensions, indicating strong performance in both structural and stylistic aspects of storytelling. This further confirms the effectiveness of the system across a range of reader expectations. Compared to the automatic evaluation results in Figure~\ref{fig:performance}(a), the trends are largely aligned, suggesting that automatic metrics provide a reasonably accurate reflection of human judgments. Nonetheless, some discrepancies are observed, as human evaluators show a stronger preference for IBSEN over GPT-4o and DeepSeek-V3. This highlights the possibility that simulation-based narratives may offer human-relatable qualities that are not fully captured by automated metrics.

\subsection{Simulation Duration Study}

\begin{figure}[t]
  \centering
  \includegraphics[width=\columnwidth]{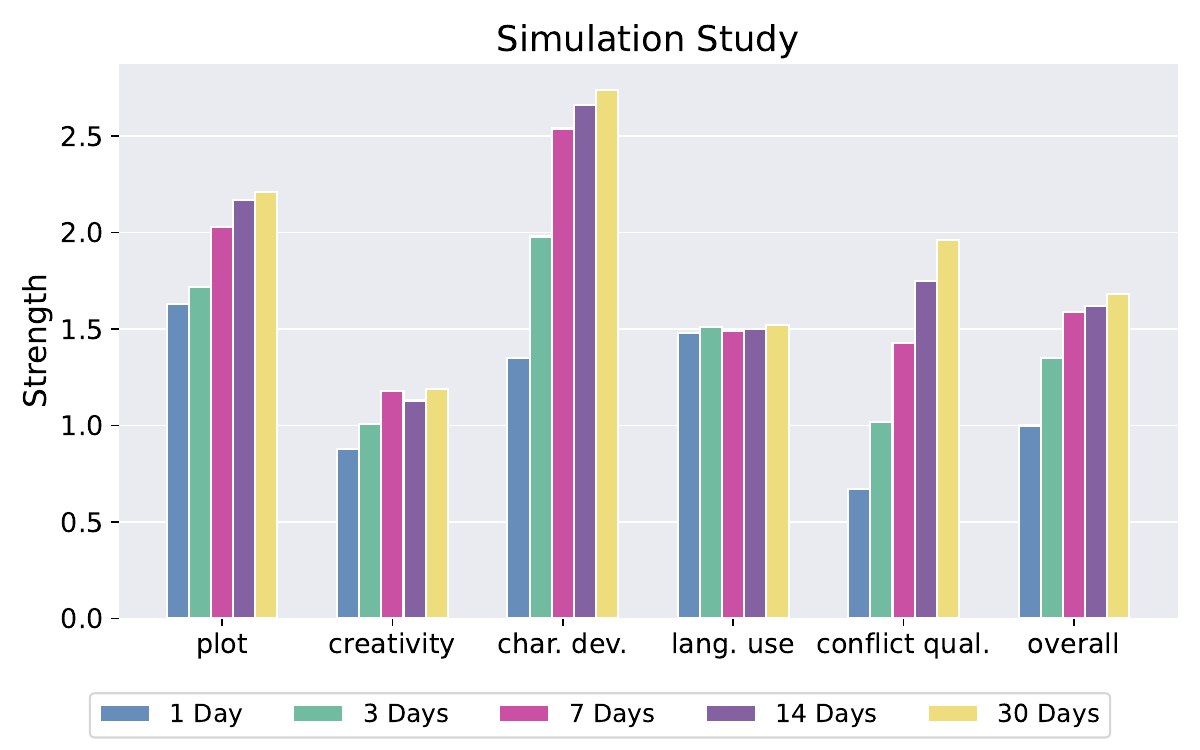}
  \caption{Effect of simulation duration on story generation performance.}
  \label{fig:simulation_study}
\end{figure}

We investigate how varying simulation durations (1, 3, 7, 14, and 30 days) affect story quality, while keeping story length fixed at around 12,000 words. As shown in Figure~\ref{fig:simulation_study}, not all metrics benefit equally from longer simulations.

\textbf{Plot}, \textbf{Creativity}, and \textbf{Language Use} show minimal variation across durations, likely due to the strong guidance of the Storyteller Agent and the language model's inherent capabilities. In contrast, \textbf{Character Development} and \textbf{Conflict Quality} improve with longer simulations, as extended interactions allow for deeper character arcs and more complex tensions. The \textbf{Overall} metric improves significantly from 1 to 7 days, but shows diminishing returns beyond that. Notably, token usage roughly doubles with each duration increase, while quality gains plateau. Excessive events in longer simulations may even burden the model's ability to synthesize information effectively.

Overall, a 7-day simulation strikes a practical balance between quality and efficiency, making it a suitable default under current model constraints.

\subsection{Ablation Study}

\begin{figure}[t]
  \centering
  \includegraphics[width=\columnwidth]{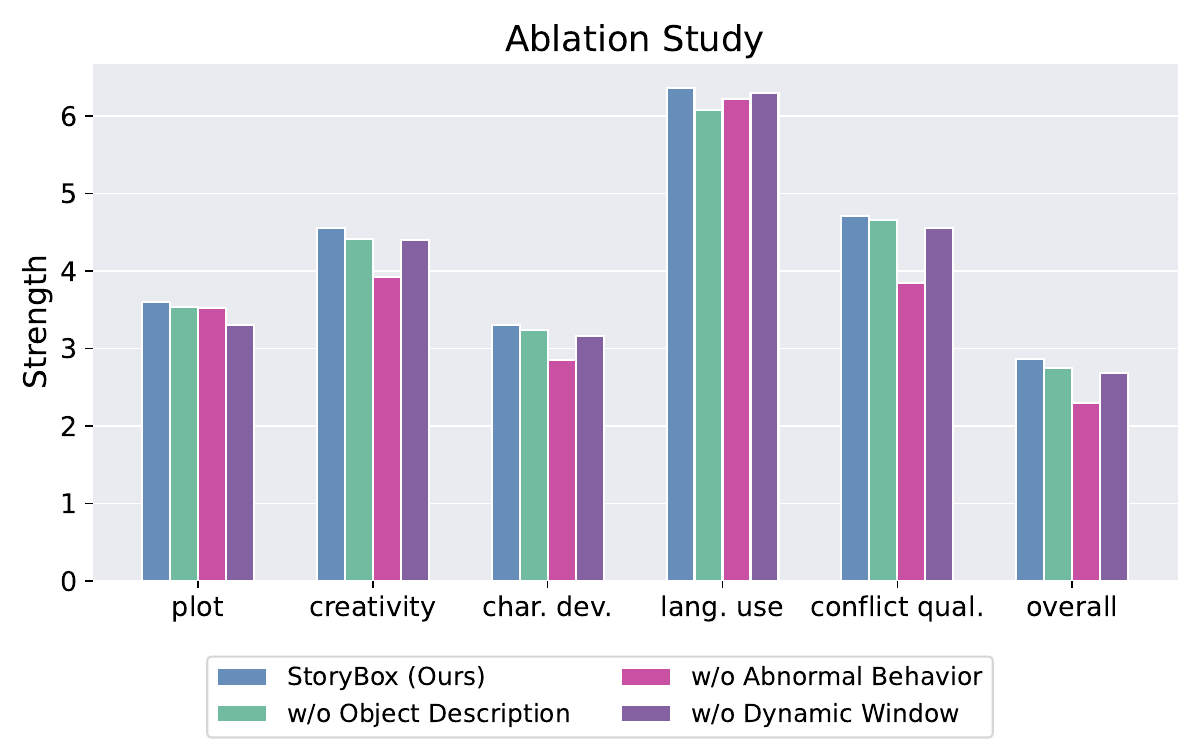}
  \caption{Performance comparison of StoryBox without different components.}
  \label{fig:ablation_study}
\end{figure}

We conduct an ablation study to assess the impact of three components: (1) object descriptions, (2) random abnormal behaviors, and the (3) dynamic context window. As shown in Figure~\ref{fig:ablation_study}, removing any component leads to a performance drop, though to varying degrees.

Excluding \textbf{object descriptions} harms \textbf{Language Use}, as environmental details help the Storyteller Agent produce richer and more grounded narratives. Removing \textbf{random abnormal behaviors} causes the sharpest decline, particularly in \textbf{Creativity}, \textbf{Character Development}, and \textbf{Conflict Quality}, highlighting the importance of unpredictability in driving dynamic stories. Disabling the \textbf{dynamic context window} reduces \textbf{Plot} quality, as it limits the model's ability to maintain coherence across longer narrative arcs. Overall, each component plays a distinct role, and their combination is key to StoryBox's effectiveness.

\section{Limitations}

While our multi-agent sandbox-based approach shows strong potential, it still has limitations. First, the simulation currently proceeds sequentially, with each character acting one at a time, which slows down the overall process. Parallelizing the simulation could improve efficiency, but managing interdependent character behaviors remains a challenge. For instance, one character's actions may directly affect others, and naive parallelization could introduce inconsistencies. Second, evaluating story quality is still a difficult problem. Although we use both human and automatic evaluations, human judgment is costly and time-consuming. More accurate and scalable automated metrics are needed to better approximate human preferences and enable broader application. These issues represent important directions for future work.

\section{Conclusion}

In this paper, we introduced StoryBox, a novel approach for long-form story generation using multi-agent simulations. Our experiments show that StoryBox outperforms existing methods on many metrics. Despite challenges in evaluating story quality, both automatic and human evaluations confirm its effectiveness in generating engaging and coherent stories.

\section*{Acknowledgements}

The authors are supported by the National Key Research and Development Program of China (Grant No. 2020YFA0712500). The authors would also like to thank National Supercomputer Center in Guangzhou for providing high performance computational resources.

\bibliography{ref}

\appendix
\clearpage

\section{Environment Modeling Details}
\label{sec:env_model_details}

\begin{figure}[t]
  \centering
  \includegraphics[width=\columnwidth]{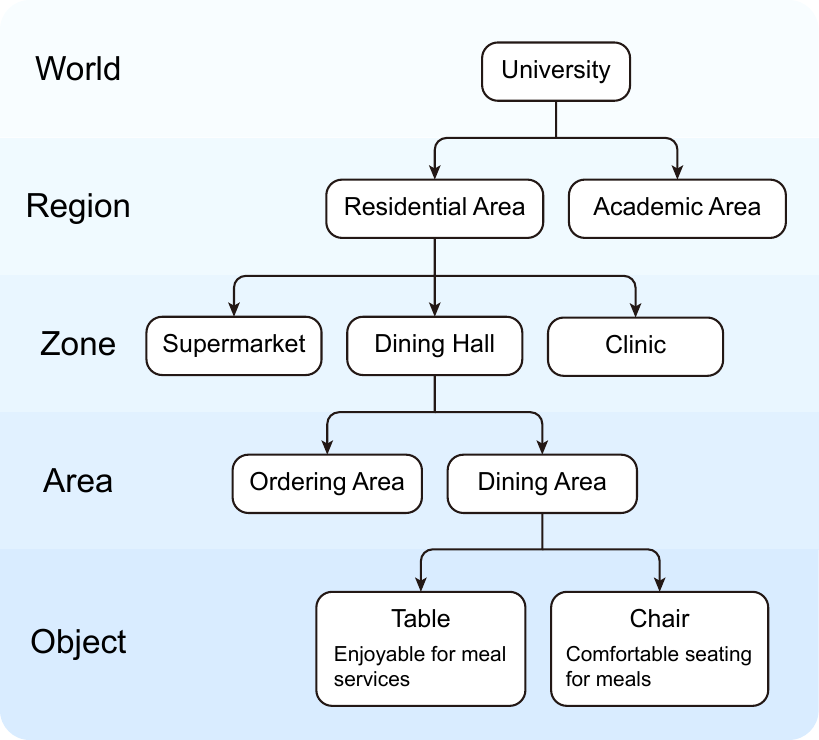}
  \caption{Overview of environment modeling using a tree-like structure with five hierarchical levels, enabling a flexible and expansive environment.}
  \label{fig:environment}
\end{figure}

As shown in Figure~\ref{fig:environment}, we divide the environment into five hierarchical levels: World, Region, Zone, Area, and Object, progressing from the broadest to the most specific. At the final level, Object, each element includes a description to capture its unique characteristics and role within the environment. Additionally, every level in this hierarchy can be assigned a description, although, for simplicity, we omit the description attributes for all levels except Object in the figure.

This hierarchical structure allows characters to perceive their environment in a flexible and context-sensitive manner. Each character can be aware of their current location as well as the objects and features within that location. By modeling the environment this way, we create a more dynamic, expansive setting for agent interactions. The environment can scale beyond the limitations imposed by tile-based systems, which typically constrain the modeling to a limited simulated space, as seen in Generative Agents. This tree-like model offers flexibility, allowing for a larger, more detailed environment that adapts to the unfolding story.

\section{Dataset Details}
\label{sec:dataset_details}

Table~\ref{tab:dataset_20_stories} presents a collection of 20 stories, each defined by a premise, setting, and characters. Each story includes the following key elements:

\begin{itemize}
  \item \textbf{Premise:} It provides a concise description of the story's core plot, setting the stage for the reader to understand the main story. This section typically outlines the primary event or challenge in the story, whether it's an adventure, a mystery, or a personal transformation. The premise introduces the main conflict or task that the protagonist must address, giving the story its driving force.
  
  \item \textbf{Setting:} It describes the time, place, and environment in which the story takes place. Settings can vary widely, from futuristic cities to ancient ruins, from dystopian societies to magical realms. The setting plays an important role in shaping the story, influencing the characters' actions and the overall tone of the story.
  
  \item \textbf{Characters:} The characters section lists the key players in each story, providing brief descriptions of their backgrounds. The characters are the driving force of the plot, and their development is crucial to the unfolding of the story. Each character's traits, relationships, and actions contribute to the story's progression and the resolution of the conflict. By understanding the characters' roles, readers can gain insight into the emotional depth and thematic elements of the story.
\end{itemize}

\onecolumn

\begin{small}
  \begin{longtable}{p{\textwidth}}
    \toprule
    \multicolumn{1}{c}{\textbf{Story 1}} \\
    \midrule
    \texttt{\textbf{[Premise]}} \\
    \texttt{After a strange phenomenon causes time to freeze for everyone except for a small group of individuals, a young scientist named Claire must find a way to reverse the event before she loses her sanity.} \\\\
    \texttt{\textbf{[Setting]}} \\
    \texttt{The story takes place in a modern-day city that has suddenly fallen into an eerie state of paralysis, with the world frozen in place.} \\\\
    \texttt{\textbf{[Characters]}} \\
    \texttt{Claire Matthews: A brilliant but socially awkward physicist in her early 30s.} \\
    \texttt{Dr. Harold Reed: An older scientist and Claire's mentor.} \\
    \texttt{Tommy Harris: A troubled teenager who sees the event as a chance to escape his problems.} \\
    \texttt{Sophia Lutz: A police officer trying to maintain order in the chaos.} \\
    \texttt{Chris Tanaka: A tech expert who believes the phenomenon is a computer glitch.} \\
    \texttt{Maya Harrison: A woman who was in the middle of an argument with her partner when time froze.} \\
    \midrule

    \multicolumn{1}{c}{\textbf{Story 2}} \\
    \midrule
    \texttt{\textbf{[Premise]}} \\
    \texttt{In a dystopian future where all art is illegal, a rebellious painter named Felix risks his life to create forbidden masterpieces in secret.} \\\\
    \texttt{\textbf{[Setting]}} \\
    \texttt{A totalitarian society in the near future where government surveillance is constant, and all forms of art are outlawed.} \\\\
    \texttt{\textbf{[Characters]}} \\
    \texttt{Felix Hartman: A young and passionate artist who defies the oppressive regime.} \\
    \texttt{Lena Stark: Felix's childhood friend, a government enforcer tasked with tracking down dissenters.} \\
    \texttt{Commander Eriksson: The ruthless leader of the government's art censorship division.} \\
    \texttt{Jasper Fox: An underground art dealer who helps Felix distribute his works.} \\
    \texttt{Sarah Hunter: A former art critic turned rebel who now works with Felix.} \\
    \midrule
  
    \multicolumn{1}{c}{\textbf{Story 3}} \\
    \midrule
    \texttt{\textbf{[Premise]}} \\
    \texttt{A struggling musician, Jordan, discovers a mysterious old piano in an abandoned mansion, only to realize that the piano has the power to transport him to alternate realities.} \\\\
    \texttt{\textbf{[Setting]}} \\
    \texttt{The story is set in a small town, with the mansion located at its outskirts, surrounded by dense woods.} \\\\
    \texttt{\textbf{[Characters]}} \\
    \texttt{Jordan Hayes: A down-on-his-luck musician in his late twenties.} \\
    \texttt{Evelyn Moore: A local historian who knows the mansion's dark past.} \\
    \texttt{Nathan Green: Jordan's childhood friend who believes the piano holds a dangerous secret.} \\
    \texttt{Mrs. Montgomery: An eccentric old woman who once lived in the mansion.} \\
    \midrule

    \multicolumn{1}{c}{\textbf{Story 4}} \\
    \midrule
    \texttt{\textbf{[Premise]}} \\
    \texttt{During a research expedition in the Arctic, a team of scientists discovers a hidden alien artifact that begins to influence their minds in unexpected ways.} \\\\
    \texttt{\textbf{[Setting]}} \\
    \texttt{The Arctic wilderness, an isolated research station miles from civilization, where snowstorms are frequent.} \\\\
    \texttt{\textbf{[Characters]}} \\
    \texttt{Dr. Emily Reynolds: A lead scientist who specializes in extraterrestrial artifacts.} \\
    \texttt{Dr. Ian McCallister: A skeptical geologist who dismisses the artifact as a hoax.} \\
    \texttt{Lena Novak: A biologist with a deep knowledge of Arctic ecosystems.} \\
    \texttt{James Archer: A security officer who is wary of the artifact's strange effects.} \\
    \texttt{Eliot White: A researcher obsessed with uncovering the artifact's true origins.} \\
    \midrule
  
    \multicolumn{1}{c}{\textbf{Story 5}} \\
    \midrule
    \texttt{\textbf{[Premise]}} \\
    \texttt{A young woman named Ava discovers that her family's ancestral home is cursed, and she must unravel its dark secrets before the curse consumes her entire bloodline.} \\\\
    \texttt{\textbf{[Setting]}} \\
    \texttt{An ancient, decaying mansion in a remote village surrounded by mist and dense forests.} \\\\
    \texttt{\textbf{[Characters]}} \\
    \texttt{Ava Lawrence: A determined and intelligent woman in her mid-twenties who inherits the family estate.} \\
    \texttt{Edward Lawrence: Ava's estranged father, who disappeared years ago under mysterious circumstances.} \\
    \texttt{Grace Thornwell: A local historian who warns Ava about the mansion's dark past.} \\
    \texttt{Jared Wilson: A young journalist who investigates the curse and becomes romantically involved with Ava.} \\
    \texttt{The Specter: A mysterious figure that haunts the mansion and seems to control its curse.} \\
    \midrule
  
    \multicolumn{1}{c}{\textbf{Story 6}} \\
    \midrule
    \texttt{\textbf{[Premise]}} \\
    \texttt{A group of strangers wake up to find themselves trapped in a massive underground maze, where they must rely on each other to survive and escape while also uncovering their shared past.} \\\\
    \texttt{\textbf{[Setting]}} \\
    \texttt{A high-tech underground facility with a sprawling maze that seems to change its structure every few hours.} \\\\
    \texttt{\textbf{[Characters]}} \\
    \texttt{Rachel Turner: A resourceful but emotionally scarred woman who used to be a military strategist.} \\
    \texttt{David Brown: A kind-hearted medical student who wants to keep the group alive.} \\
    \texttt{Victor Chang: A mysterious, seemingly aloof man who has a hidden agenda.} \\
    \texttt{Anna Schwartz: A tech expert with knowledge of the maze's design.} \\
    \texttt{Jason Miller: A former prison guard who is accustomed to dealing with dangerous people.} \\
    \midrule
  
    \multicolumn{1}{c}{\textbf{Story 7}} \\
    \midrule
    \texttt{\textbf{[Premise]}} \\
    \texttt{In a magical kingdom where elements are controlled by wizards, a young orphan named Finn discovers he has the power to control a rare and forbidden element, chaos, and must learn to control it before it destroys everything.} \\\\
    \texttt{\textbf{[Setting]}} \\
    \texttt{A fantastical kingdom with floating castles, enchanted forests, and dangerous creatures.} \\\\
    \texttt{\textbf{[Characters]}} \\
    \texttt{Finn Colton: A brave and curious 16-year-old orphan who discovers his power.} \\
    \texttt{Master Alden: A wise and mysterious wizard who trains Finn in the ways of elemental magic.} \\
    \texttt{Lira Ardent: A skilled fire mage who becomes Finn's closest ally.} \\
    \texttt{King Roderick: The ruler of the kingdom, who wants to control Finn's powers for his own gain.} \\
    \texttt{Vera Duskwood: A shadowy figure who has her own dark plans for Finn's chaos magic.} \\
    \midrule
  
    \multicolumn{1}{c}{\textbf{Story 8}} \\
    \midrule
    \texttt{\textbf{[Premise]}} \\
    \texttt{A detective investigating a series of seemingly unrelated murders starts receiving cryptic messages from a mysterious informant who seems to know the truth about the crimes before they occur.} \\\\
    \texttt{\textbf{[Setting]}} \\
    \texttt{A rainy, noir-style city with narrow alleys and neon lights casting long shadows.} \\\\
    \texttt{\textbf{[Characters]}} \\
    \texttt{Detective Marcus Kane: A jaded detective in his forties, struggling with his own demons.} \\
    \texttt{Vivienne Stone: A mysterious informant who only communicates through letters and phone calls.} \\
    \texttt{Sergeant Alan Pierce: Marcus's loyal but frustrated partner who wants to solve the case by the book.} \\
    \texttt{Martha Lawson: A grieving mother whose daughter was one of the victims.} \\
    \texttt{Adrian West: A high-ranking politician with a questionable connection to the victims.} \\
    \midrule

    \multicolumn{1}{c}{\textbf{Story 9}} \\
    \midrule
    \texttt{\textbf{[Premise]}} \\
    \texttt{A young journalist named Harper stumbles upon a secret society of time travelers, and she must decide whether to join them in their fight to protect history or expose them to the world.} \\\\
    \texttt{\textbf{[Setting]}} \\
    \texttt{Modern-day New York City with hidden passageways that lead to a network of time-travel portals.} \\\\
    \texttt{\textbf{[Characters]}} \\
    \texttt{Harper Wells: An ambitious and fearless journalist who becomes entangled in time travel.} \\
    \texttt{Dorian Blackwell: A charismatic leader of the time-traveling society who has lived for centuries.} \\
    \texttt{Liam Quinn: A former history professor who is skeptical of the society's methods.} \\
    \texttt{Isla Byrne: A member of the society who specializes in technology that aids time travel.} \\
    \texttt{Agent Romanov: A government agent who is investigating the society's existence.} \\
    \midrule
  
    \multicolumn{1}{c}{\textbf{Story 10}} \\
    \midrule
    \texttt{\textbf{[Premise]}} \\
    \texttt{A detective with the ability to read minds becomes entangled in a complex case involving a missing child, a web of lies, and the darker side of his own abilities.} \\\\
    \texttt{\textbf{[Setting]}} \\
    \texttt{A city divided between wealth and poverty, with dark corners where crime festers.} \\\\
    \texttt{\textbf{[Characters]}} \\
    \texttt{Detective Leo Novak: A sharp-witted detective in his thirties who struggles with his mind-reading powers.} \\
    \texttt{Amanda Giles: The desperate mother of the missing child who is hiding her own secrets.} \\
    \texttt{Henry Cole: A criminal mastermind whose plans are often obscured by his charismatic personality.} \\
    \texttt{Jenna Harrow: Leo's ex-wife who helps him investigate the case despite their complicated past.} \\
    \texttt{Detective Rita Moon: A no-nonsense investigator who mistrusts Leo's unconventional methods.} \\
    \midrule
  
    \multicolumn{1}{c}{\textbf{Story 11}} \\
    \midrule
    \texttt{\textbf{[Premise]}} \\
    \texttt{A group of astronauts on a deep-space mission discover an ancient alien vessel that contains a mysterious substance capable of changing reality, but using it comes at a great cost.} \\\\
    \texttt{\textbf{[Setting]}} \\
    \texttt{Aboard a high-tech space station orbiting a distant, uncharted planet, with the alien vessel located on its surface.} \\\\
    \texttt{\textbf{[Characters]}} \\
    \texttt{Captain Elena Ruiz: The commanding officer of the mission, responsible for the crew's safety.} \\
    \texttt{Dr. Marcus Trent: A scientist who is fascinated by the alien technology and its potential.} \\
    \texttt{Commander Kai Chen: A pragmatic and cautious officer who is skeptical of the substance.} \\
    \texttt{Mia Sanchez: A young engineer who begins to experience strange visions after interacting with the substance.} \\
    \texttt{Zane Holt: A communications officer who is unknowingly being influenced by the reality-altering substance.} \\
    \midrule
  
    \multicolumn{1}{c}{\textbf{Story 12}} \\
    \midrule
    \texttt{\textbf{[Premise]}} \\
    \texttt{A retired private investigator, Jack, is forced to return to his old profession when his estranged daughter is kidnapped, and he is given a cryptic message from the kidnapper.} \\\\
    \texttt{\textbf{[Setting]}} \\
    \texttt{A gritty coastal city with a criminal underworld, dim-lit alleyways, and seedy nightclubs.} \\\\
    \texttt{\textbf{[Characters]}} \\
    \texttt{Jack Lawson: A hardened ex-private investigator in his mid-40s, who is trying to put his past behind him.} \\
    \texttt{Eliza Lawson: Jack's estranged daughter, a young woman who has fallen into the wrong crowd.} \\
    \texttt{Vincent Marlowe: A mysterious criminal figure who may have information about Eliza's disappearance.} \\
    \texttt{Rita Blackwood: A former associate of Jack's who has a complicated relationship with him.} \\
    \texttt{Detective Claire Moore: A determined detective who is reluctantly forced to team up with Jack.} \\
    \midrule
  
    \multicolumn{1}{c}{\textbf{Story 13}} \\
    \midrule
    \texttt{\textbf{[Premise]}} \\
    \texttt{In a small town, a group of teenagers begins to discover that their town is a gateway between dimensions, and they must stop an evil force from crossing into their world.} \\\\
    \texttt{\textbf{[Setting]}} \\
    \texttt{A picturesque but eerie small town with strange occurrences and hidden portals to other worlds.} \\\\
    \texttt{\textbf{[Characters]}} \\
    \texttt{Sammy Rivers: A brave but reluctant leader of the group of teenagers.} \\
    \texttt{Lily Walker: A sharp-witted girl with a keen sense of the supernatural.} \\
    \texttt{Ethan Hayes: A skeptic who doesn't believe in the dimensions until he experiences them firsthand.} \\
    \texttt{Mason Cruz: A quiet boy who has strange dreams that hint at the town's hidden powers.} \\
    \texttt{Mayor Grace Turner: The town's mayor, who knows more about the dimensional gateways than she lets on.} \\
    \midrule
  
    \multicolumn{1}{c}{\textbf{Story 14}} \\
    \midrule
    \texttt{\textbf{[Premise]}} \\
    \texttt{A team of archaeologists uncovers an ancient temple in the jungle that holds the key to an ancient civilization's downfall, but releasing its secrets may bring about the same fate for them.} \\\\
    \texttt{\textbf{[Setting]}} \\
    \texttt{A dense, overgrown jungle in Central America, with a mysterious, hidden temple at its heart.} \\\\
    \texttt{\textbf{[Characters]}} \\
    \texttt{Dr. Emily Hayes: A passionate archaeologist who is determined to unlock the temple's secrets.} \\
    \texttt{Dr. Lucas Donovan: A pragmatic archaeologist who is more concerned about the safety of the team.} \\
    \texttt{Carla Vasquez: A local guide who knows the legends of the temple but refuses to venture near it.} \\
    \texttt{Raj Patel: A tech expert who uncovers ancient artifacts that hint at a deadly curse.} \\
    \texttt{Kara Moore: An experienced survivalist who is skeptical of the supernatural events surrounding the temple.} \\
    \midrule
  
    \multicolumn{1}{c}{\textbf{Story 15}} \\
    \midrule
    \texttt{\textbf{[Premise]}} \\
    \texttt{A talented but jaded painter, Owen, is cursed to live in a never-ending cycle of painting the same masterpiece for eternity, unable to escape until he finds the true meaning of his art.} \\\\
    \texttt{\textbf{[Setting]}} \\
    \texttt{A small, isolated art studio located in a remote village on the edge of a cliff, overlooking a stormy sea.} \\\\
    \texttt{\textbf{[Characters]}} \\
    \texttt{Owen Price: A once-celebrated painter now trapped in a timeless, painful cycle.} \\
    \texttt{Mariana Clark: A fellow artist who helps Owen understand the deeper meaning of his work.} \\
    \texttt{Victor Sands: A mysterious stranger who may have cursed Owen to this eternal cycle.} \\
    \texttt{Hector Hayes: Owen's long-time mentor who abandoned him years ago, leading to his current predicament.} \\
    \midrule
  
    \multicolumn{1}{c}{\textbf{Story 16}} \\
    \midrule
    \texttt{\textbf{[Premise]}} \\
    \texttt{A young woman named Lara discovers that she is the heir to a hidden kingdom beneath the earth's surface, and she must navigate ancient politics and betrayals to reclaim her birthright.} \\\\
    \texttt{\textbf{[Setting]}} \\
    \texttt{A hidden, technologically advanced underground kingdom, with subterranean cities and vast caverns.} \\\\
    \texttt{\textbf{[Characters]}} \\
    \texttt{Lara Sinclair: A determined and courageous woman in her early twenties who is shocked to learn of her heritage.} \\
    \texttt{King Malcus: The enigmatic ruler of the underground kingdom who has his own interests in Lara's return.} \\
    \texttt{Jarek Voss: A charismatic rebel leader who seeks to overthrow the current regime and recruit Lara.} \\
    \texttt{Elda Starling: An ancient guardian of the underground kingdom who protects its secrets.} \\
    \texttt{Victor Denholm: A power-hungry noble who is determined to prevent Lara from claiming her birthright.} \\
    \midrule
  
    \multicolumn{1}{c}{\textbf{Story 17}} \\
    \midrule
    \texttt{\textbf{[Premise]}} \\
    \texttt{In a world where dreams can be controlled and manipulated, a group of thieves specialize in entering people's dreams to steal their deepest secrets, but when one of them begins to lose control, the dreamscape turns deadly.} \\\\
    \texttt{\textbf{[Setting]}} \\
    \texttt{A cyberpunk city in the near future, where technology has advanced to the point that dreams can be accessed and altered.} \\\\
    \texttt{\textbf{[Characters]}} \\
    \texttt{Elliot Dray: A skilled dream thief, haunted by his past and beginning to lose his grip on reality.} \\
    \texttt{Juno Vane: A brilliant but ruthless hacker who leads the team of dream thieves.} \\
    \texttt{Mila Roswell: A former psychologist turned dream thief who can navigate complex subconscious landscapes.} \\
    \texttt{Dr. Harris Lennox: A neuroscientist who develops the technology that allows people to enter and manipulate dreams.} \\
    \texttt{Darren Oakley: A mysterious figure from Elliot's past who is connected to his growing inability to control his own dreams.} \\
    \midrule
  
    \multicolumn{1}{c}{\textbf{Story 18}} \\
    \midrule
    \texttt{\textbf{[Premise]}} \\
    \texttt{A small-town librarian, Margaret, begins receiving strange letters from an anonymous person who claims to know about a hidden treasure buried beneath the town, leading her to question the history of her hometown and its secrets.} \\\\
    \texttt{\textbf{[Setting]}} \\
    \texttt{A quiet, picturesque small town with a long history of strange rumors and forgotten legends, surrounded by dense forests and mountains.} \\\\
    \texttt{\textbf{[Characters]}} \\
    \texttt{Margaret Reed: A quiet and intelligent librarian in her late thirties, curious about her town's past.} \\
    \texttt{Oliver Finch: A local historian who has dedicated his life to studying the town's folklore.} \\
    \texttt{Rachel Turner: Margaret's best friend, a skeptic who believes the letters are a hoax.} \\
    \texttt{Mayor Thomas Cole: The charming but secretive mayor, who is suspicious of Margaret's investigation.} \\
    \texttt{Eliot Hawke: An eccentric treasure hunter who arrives in town, claiming to know the true location of the treasure.} \\
    \midrule

    \multicolumn{1}{c}{\textbf{Story 19}} \\
    \midrule
    \texttt{\textbf{[Premise]}} \\
    \texttt{A group of outcasts, each with a personal vendetta against a corrupt corporation, band together to carry out a heist that will expose the company's darkest secrets to the world, but they soon realize that the corporation's power goes far beyond their expectations.} \\\\
    \texttt{\textbf{[Setting]}} \\
    \texttt{A futuristic metropolis controlled by a powerful and shadowy corporation that manipulates both the government and the media.} \\\\
    \texttt{\textbf{[Characters]}} \\
    \texttt{Cassie Parker: A former corporate insider turned hacker, who seeks revenge on the company that ruined her career.} \\
    \texttt{Jared Cross: A former soldier with a deep hatred for the corporation after they betrayed his team.} \\
    \texttt{Sophia Nash: A tech expert who has been living off-the-grid, hiding from the corporation's surveillance.} \\
    \texttt{Dante Moore: A smooth-talking con artist who uses his charm to manipulate others for the cause.} \\
    \texttt{Director Lyle Hayes: The ruthless CEO of the corporation, whose crimes are hidden by layers of influence and control.} \\
    \midrule
  
    \multicolumn{1}{c}{\textbf{Story 20}} \\
    \midrule
    \texttt{\textbf{[Premise]}} \\
    \texttt{A young archaeologist, Theo, discovers a hidden cave system filled with ancient drawings that seem to predict the future. As he unravels the mystery, he is drawn into a dangerous race against time to prevent a cataclysmic event from occurring.} \\\\
    \texttt{\textbf{[Setting]}} \\
    \texttt{A remote desert region with ancient caves, hidden temples, and a long-forgotten civilization buried beneath the sand.} \\\\
    \texttt{\textbf{[Characters]}} \\
    \texttt{Theo Carter: A passionate and idealistic archaeologist who stumbles upon the cave and its secrets.} \\
    \texttt{Dr. Amina Zafir: An experienced archaeologist and Theo's mentor, who is more skeptical of the cave's significance.} \\
    \texttt{Rafael Morales: A local guide with knowledge of the desert's legends, who becomes Theo's reluctant ally.} \\
    \texttt{Commander Isabella Grant: A military officer who is tasked with investigating the caves and the potential threat they pose.} \\
    \texttt{Elder Karim: A wise figure from a nearby village who believes that the ancient drawings hold the key to the world's survival.} \\
    \bottomrule

    \caption{Detailed description of the 20 stories in the dataset, including their premise, setting, and characters.}
    \label{tab:dataset_20_stories}
  \end{longtable}
\end{small}

\twocolumn

\section{Metric Details}
\label{sec:metric_details}

This section provides detailed descriptions of the evaluation metrics used for long-form story generation, including both human evaluation metrics and automatic evaluation metrics.

\subsection{Human Evaluation Metrics}

\begin{itemize}
  \item \textbf{Plot:} This metric evaluates the structural integrity and coherence of the narrative. A well-executed plot presents a clearly defined beginning, middle, and end, with events that unfold in a logical and engaging sequence. Strong plots maintain a clear narrative direction through well-paced turning points and developments that align with the story's themes and internal logic. Avoidance of plot holes, abrupt transitions, or confusing sequences further contributes to the clarity and momentum of the story.

  \item \textbf{Creativity:} This metric measures the originality and imaginative strength of the narrative. A creative story introduces fresh ideas, vivid characters, and emotionally resonant themes while steering clear of clichés, stereotypes, and formulaic tropes. The ability to craft a distinctive world and storyline without leaning heavily on the prompt reflects strong creative intent. When characters, events, and settings feel unexpected yet meaningful, the story gains a sense of uniqueness and depth.
  
  \item \textbf{Character Development:} This metric evaluates the depth and progression of characters within the story. Strong character development features multi-dimensional individuals whose motivations, behaviors, and decisions are clearly defined and remain consistent with their established personas. Characters who exhibit noticeable growth, transformation, or internal struggle across the narrative demonstrate a compelling arc that enhances emotional resonance. Rich backstories and emotional complexity further reinforce the believability and depth of the characters, making them more engaging and authentic.
  
  \item \textbf{Language Use:} This metric evaluates the richness, precision, and stylistic impact of the language. Effective language use is marked by varied sentence structures, purposeful word choices, and the use of literary devices such as metaphor, imagery, and descriptive detail to enhance tone and atmosphere. Strong narratives employ language to vividly portray settings, immersing the reader in the physical and emotional landscape of the story. A compelling narrative voice avoids repetition or flat expression, using language as a tool to deepen characterization and sustain reader engagement.
  
  \item \textbf{Conflict Quality:} This metric assesses the presence, intensity, and narrative function of conflict within the story. High-quality conflict drives the plot forward, deepens character arcs, and highlights thematic elements through meaningful tension and stakes. Effective conflict is not only believable and engaging, but also complex in its emotional or moral dimensions, providing opportunities for growth, confrontation, and resolution. The way conflict shapes the overall trajectory of the narrative reflects the story's structural and emotional depth.
  
\end{itemize}

\subsection{Automatic Evaluation Metrics}

Our automatic evaluation metrics consist of two components. First, we adopt the same dimensions used in human evaluation, but apply them through structured assessments conducted by a large language model. This enables a systematic and efficient approximation of human judgments.

Second, we extend the evaluation with additional metrics tailored to our task. In particular, we introduce a specialized metric for multi-agent simulation-based storytelling, which we call Character Behavior Consistency, as well as a simple statistical indicator, Average Word Count, which provides insight into the overall story length. The detailed definitions are as follows:

\begin{itemize}
  \item \textbf{Character Behavior Consistency:} A specialized metric for multi-agent simulation-based story generation, this measures whether characters' actions align with their predefined persona settings (Persona Scratch Information). Characters should act consistently with their attributes, goals, and motivations, ensuring believability and coherence in the story.
  
  \item \textbf{Average Word Count:} This metric calculates the average number of words per generated story. While basic, it provides an indication of the story's overall length, which is crucial in long-form generation. Longer stories typically allow for more extensive character development and greater plot complexity.
\end{itemize}

\section{Human Evaluation Details}
\label{sec:human_eval_details}

\begin{figure*}[htbp]
  \centering
  \includegraphics[width=\textwidth]{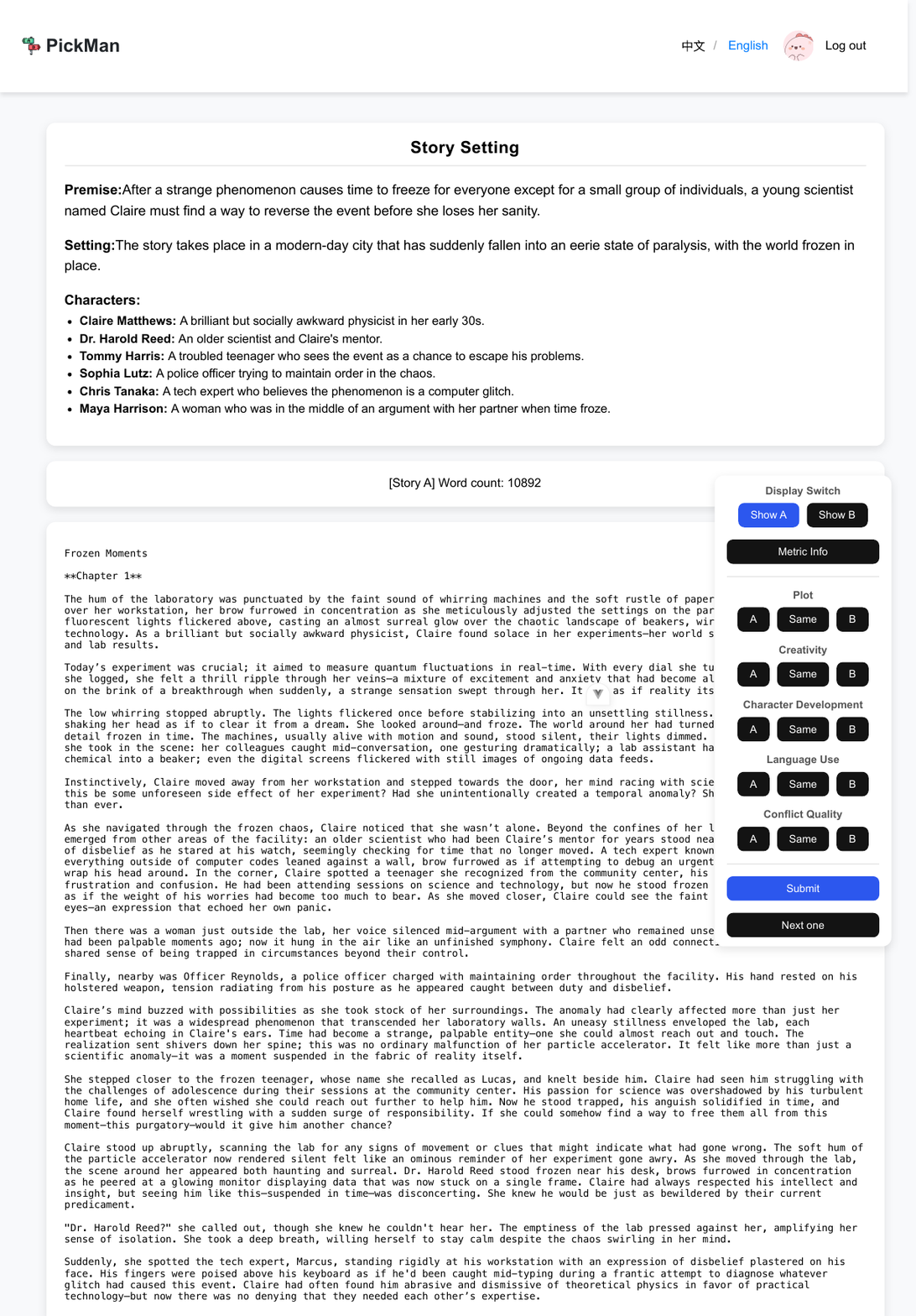}
  \caption{Overview of the human evaluation system.}
  \label{fig:human-eval-system}
\end{figure*}

To facilitate consistent and efficient story evaluation, we develop a custom web-based platform tailored to the needs of our human evaluation task. An overview of the platform interface is shown in Figure~\ref{fig:human-eval-system}. To accommodate a diverse group of annotators, the platform supports both Chinese and English language modes. Access to the system is restricted to verified annotators, who must log in before participating in the evaluation process to ensure traceability and avoid duplicate submissions.

Each evaluation instance is based on a specific story setting, under which two stories generated by different methods are presented as Story A and Story B. For each story, the interface displays the complete text along with the word count, enabling annotators to consider both narrative content and overall length. Annotators can switch between Story A and Story B using the controls on the right-hand panel, where they can also view detailed explanations of the evaluation criteria. For each criterion, annotators are asked to indicate which story they prefer, or to mark them as equivalent if no preference can be determined. Once all selections are made, the annotator submits the evaluation and is presented with the next story pair. The backend system maintains a record of which story pairs have been evaluated by each annotator, thereby preventing duplicate evaluations.

\section{Automatic Evaluation Details}
\label{sec:auto_eval_details}

To conduct automatic evaluation, we follow a prompting strategy similar to that used in Agents' Room, with carefully structured templates that guide the model to emulate human annotation. An example of the prompt template used in our evaluation is shown in Table~\ref{tab:auto_eval_prompt_template}. The prompt includes detailed definitions of the five evaluation dimensions to ensure consistency with human evaluation standards. The LLM is instructed to assess each story pair along these dimensions and provide a final preference for each, enabling us to systematically collect model-based judgments that are directly comparable to human annotations.

\begin{table*}[]
  \small
  \centering
  \begin{tabularx}{\textwidth}{X}
    \toprule
    \texttt{You will conduct a side-by-side evaluation. You will be given two system-generated stories. Your task is to compare the two stories and determine which one is better based on the following dimensions:} \\\\

    \texttt{\textbf{Plot:} This metric evaluates the structural integrity and coherence of the narrative. A well-executed plot presents a clearly defined beginning, middle, and end, with events that unfold in a logical and engaging sequence. Strong plots maintain a clear narrative direction through well-paced turning points and developments that align with the story's themes and internal logic. Avoidance of plot holes, abrupt transitions, or confusing sequences further contributes to the clarity and momentum of the story.} \\\\

    \texttt{\textbf{Creativity:} This metric measures the originality and imaginative strength of the narrative. A creative story introduces fresh ideas, vivid characters, and emotionally resonant themes while steering clear of clichés, stereotypes, and formulaic tropes. The ability to craft a distinctive world and storyline without leaning heavily on the prompt reflects strong creative intent. When characters, events, and settings feel unexpected yet meaningful, the story gains a sense of uniqueness and depth.} \\\\
    
    \texttt{\textbf{Character Development:} This metric evaluates the depth and progression of characters within the story. Strong character development features multi-dimensional individuals whose motivations, behaviors, and decisions are clearly defined and remain consistent with their established personas. Characters who exhibit noticeable growth, transformation, or internal struggle across the narrative demonstrate a compelling arc that enhances emotional resonance. Rich backstories and emotional complexity further reinforce the believability and depth of the characters, making them more engaging and authentic.} \\\\
    
    \texttt{\textbf{Language Use:} This metric evaluates the richness, precision, and stylistic impact of the language. Effective language use is marked by varied sentence structures, purposeful word choices, and the use of literary devices such as metaphor, imagery, and descriptive detail to enhance tone and atmosphere. Strong narratives employ language to vividly portray settings, immersing the reader in the physical and emotional landscape of the story. A compelling narrative voice avoids repetition or flat expression, using language as a tool to deepen characterization and sustain reader engagement.} \\\\

    \texttt{\textbf{Conflict Quality:} This metric assesses the presence, intensity, and narrative function of conflict within the story. High-quality conflict drives the plot forward, deepens character arcs, and highlights thematic elements through meaningful tension and stakes. Effective conflict is not only believable and engaging, but also complex in its emotional or moral dimensions, providing opportunities for growth, confrontation, and resolution. The way conflict shapes the overall trajectory of the narrative reflects the story's structural and emotional depth.} \\\\

    \texttt{Provide a detailed assessment of the two stories in terms of these five dimensions. Conclude your assessment with scores for each dimension using the template below. Do not add any emphasis, such as bold and italics, on your assessment.} \\

    \midrule
    \midrule
    \texttt{\textbf{[Story A]} <Story A>} \\\\
    \texttt{\textbf{[Story B]} <Story B>} \\
    
    \midrule
    \midrule
    \texttt{The better story for each dimension is:} \\
    \texttt{Plot: [A or B or Same]} \\
    \texttt{Creativity: [A or B or Same]} \\
    \texttt{Character Development: [A or B or Same]} \\
    \texttt{Language Use: [A or B or Same]} \\
    \texttt{Conflict Quality: [A or B or Same]} \\

    \bottomrule
  \end{tabularx}
  \caption{Prompt template used in automatic evaluation. The LLM receives two generated stories along with detailed definitions of each evaluation dimension, and is asked to compare them accordingly. Story content is omitted here for brevity.}
  \label{tab:auto_eval_prompt_template}
\end{table*}

In addition to standard qualitative metrics, we further evaluate the Character Behavior Consistency of stories generated under the multi-agent simulation framework. This metric is specifically designed to assess whether characters act in accordance with their predefined persona attributes, referred to as the Identity Stable Set (ISS). To facilitate this evaluation, we construct a prompt that guides the LLM to analyze each character's actions, decisions, and dialogue for alignment with their ISS, as shown in Table~\ref{tab:auto_eval_prompt_template_char_behav_cons}. The model is asked to rate consistency on a 0–10 scale, based on alignment, believability, and absence of contradictions.

For the statistical metric Average Word Count, we compute the average number of words per story by simply splitting text based on whitespace.

\begin{table*}[]
  \small
  \centering
  \begin{tabularx}{\textwidth}{X}
    \toprule
    \texttt{You are an expert in story evaluation and character development analysis. Your task is to assess the consistency of character behavior in a story. Consistency refers to whether the characters' actions, decisions, and dialogue align with their established traits, background, and personality as described in their Identity Stable Set.} \\

    \midrule

    \texttt{Here are the provided character descriptions (ISS) and the story for evaluation:} \\\\

    \texttt{Character Descriptions (ISS): <Character Descriptions (ISS)>} \\\\

    \texttt{Story: <Story>} \\

    \midrule

    \texttt{Evaluation Task:} \\\\

    \texttt{Behavior Consistency Analysis: Assess the story based on the following criteria:} \\\\

    \texttt{1. Alignment with ISS: Do the characters' actions, decisions, and speech align with their ISS descriptions (e.g., traits, lifestyle, and current state)?} \\

    \texttt{2. Believability: Are the characters' behaviors logical and consistent within the story's context and their established traits?} \\

    \texttt{3. Avoidance of Contradictions: Are there any moments where characters behave in a way that directly contradicts their ISS?} \\

    \midrule

    \texttt{Provide a score on a scale of 0 to 10, where:} \\

    \texttt{0 means "The characters are completely inconsistent with their ISS descriptions."} \\

    \texttt{10 means "The characters are perfectly consistent with their ISS descriptions."} \\

    \bottomrule
  \end{tabularx}
  \caption{Prompt template used for evaluating Character Behavior Consistency in multi-agent simulation-based stories. The LLM assesses whether characters' actions and dialogue are coherent with their predefined persona attributes, using a structured rubric and a 0-10 scoring scale.}
  \label{tab:auto_eval_prompt_template_char_behav_cons}
\end{table*}

\section{Sandbox Initialization}
\label{sec:sandbox_initialization}

Building a sandbox based on a multi-agent simulation requires proper initialization. The initialization process relies on the relevant settings from the dataset's stories, including the setup of sandbox characters and the sandbox environment. In this section, we provide an example of how the sandbox is initialized based on the story settings from the dataset. This example is taken from ``Story 1'' in Table~\ref{tab:dataset_20_stories}. The sandbox initialization for all stories can be found in our project.

\subsection{Character Setup}

The setup of sandbox characters is derived from the understanding of the relevant story settings. It primarily involves establishing the ``Persona Scratch Information'', as outlined in Figure~\ref{fig:framework}, which includes details such as names, ages, and other relevant information. Based on the description in ``Story 1'', we utilize \texttt{GPT-4o mini} to generate the corresponding character setups, as shown in Table~\ref{tab:story_1_character_setup}. The same approach is applied to other stories.

\subsection{Environment Setup}

The sandbox environment setup is similarly based on the understanding of the relevant story settings. The goal is to construct an environment with hierarchical relationships, as depicted in Figure~\ref{fig:environment}. In this case, drawing from the details in ``Story 1'', we use \texttt{GPT-4o mini} to generate the corresponding environment setup in the form of a YAML file, as shown in Table~\ref{tab:story_1_environment_setup}. This method is also applied to other stories in the dataset.

\onecolumn
\begin{small}
  \begin{longtable}{p{\textwidth}}
    \toprule
    \multicolumn{1}{c}{\textbf{Story 1}} \\
    \midrule
    \texttt{\textbf{[Premise]}} \\
    \texttt{After a strange phenomenon causes time to freeze for everyone except for a small group of individuals, a young scientist named Claire must find a way to reverse the event before she loses her sanity.} \\\\
    \texttt{\textbf{[Setting]}} \\
    \texttt{The story takes place in a modern-day city that has suddenly fallen into an eerie state of paralysis, with the world frozen in place.} \\\\
    \texttt{\textbf{[Characters]}} \\
    \texttt{Claire Matthews: A brilliant but socially awkward physicist in her early 30s.} \\
    \texttt{Dr. Harold Reed: An older scientist and Claire's mentor.} \\
    \texttt{Tommy Harris: A troubled teenager who sees the event as a chance to escape his problems.} \\
    \texttt{Sophia Lutz: A police officer trying to maintain order in the chaos.} \\
    \texttt{Chris Tanaka: A tech expert who believes the phenomenon is a computer glitch.} \\
    \texttt{Maya Harrison: A woman who was in the middle of an argument with her partner when time froze.} \\
    \midrule

    \multicolumn{1}{c}{\textbf{Character 1: Claire Matthews}} \\
    \midrule
    \texttt{\textbf{Name:} Claire Matthews} \\
    \texttt{\textbf{Age:} 32} \\
    \texttt{\textbf{Innate:} Brilliant, analytical, introverted} \\
    \texttt{\textbf{Learned:} Claire Matthews is a physicist with a specialization in quantum mechanics and temporal phenomena. She is highly regarded in the scientific community for her innovative research but struggles with social interactions and often immerses herself in her work to avoid personal connections.} \\
    \texttt{\textbf{Currently:} Claire Matthews is working tirelessly in a makeshift laboratory to understand the mysterious phenomenon that has frozen time. She is developing theories and conducting experiments to find a way to reverse the event, driven by a fear of isolation and a desire to restore normalcy.} \\
    \texttt{\textbf{Lifestyle:} Claire's days are now consumed by research. She works from dawn till midnight, breaking only for short meals. Her life has become a cycle of hypothesis, testing, and analysis, with minimal contact with the other individuals unaffected by the phenomenon.} \\
    \texttt{\textbf{Living Area:} Frozen City:City Center:Highland Apartments:Room 704} \\
    \texttt{\textbf{Daily Plan Requirement:}} \\
    \texttt{1. Analyze the frozen state phenomenon} \\
    \texttt{2. Conduct experiments on temporal mechanics} \\
    \texttt{3. Document findings} \\
    \texttt{4. Collaborate with Dr. Reed} \\
    \\
    \midrule
    
    \multicolumn{1}{c}{\textbf{Character 2: Dr. Harold Reed}} \\
    \midrule
    \texttt{\textbf{Name:} Dr. Harold Reed} \\
    \texttt{\textbf{Age:} 68} \\
    \texttt{\textbf{Innate:} Wise, Patient, Methodical} \\
    \texttt{\textbf{Learned:} Dr. Harold Reed is a retired physicist and former professor who has mentored many young scientists, including Claire. His expertise in theoretical physics makes him an invaluable resource in understanding the current crisis. He brings a calm, guiding presence to the chaotic situation.} \\
    \texttt{\textbf{Currently:} Dr. Harold Reed is assisting Claire with her research, offering insights and reviewing her work. He spends his time pouring over old research papers and theoretical models that might provide a clue to the current phenomenon.} \\
    \texttt{\textbf{Lifestyle:} Dr. Reed's routine is now centered around supporting Claire's efforts. He starts his day with a thorough review of scientific literature, followed by long discussions with Claire. He takes regular breaks for tea and reflection, often advising others on staying calm.} \\
    \texttt{\textbf{Living Area:} Frozen City:Suburbs:Elmwood House:Unit 12} \\
    \texttt{\textbf{Daily Plan Requirement:}} \\
    \texttt{1. Review Claire's experiments} \\
    \texttt{2. Research temporal theories} \\
    \texttt{3. Provide mentorship} \\
    \texttt{4. Maintain morale} \\
    \\
    \midrule

    \multicolumn{1}{c}{\textbf{Character 3: Tommy Harris}} \\
    \midrule
    \texttt{\textbf{Name:} Tommy Harris} \\
    \texttt{\textbf{Age:} 17} \\
    \texttt{\textbf{Innate:} Rebellious, Resourceful, Impulsive} \\
    \texttt{\textbf{Learned:} Tommy Harris has had a troubled life, facing family issues and academic struggles. He is a street-smart teenager who has learned to fend for himself. The frozen world presents him with an opportunity to escape his past and redefine himself.} \\
    \texttt{\textbf{Currently:} Tommy is exploring the frozen city, scavenging supplies, and looking for ways to use the situation to his advantage. He is also curious about the phenomenon and occasionally assists Claire and the others in practical tasks.} \\
    \texttt{\textbf{Lifestyle:} Tommy's day revolves around exploring new parts of the city, collecting items he finds valuable, and occasionally checking in with the group for food or shelter. He has a makeshift base in an abandoned store where he feels safe.} \\
    \texttt{\textbf{Living Area:} Frozen City:City Center:Abandoned Warehouse:Room 3} \\
    \texttt{\textbf{Daily Plan Requirement:}} \\
    \texttt{1. Scavenge supplies} \\
    \texttt{2. Explore the city} \\
    \texttt{3. Avoid danger} \\
    \texttt{4. Assist Claire occasionally} \\
    \\
    \midrule
    
    \multicolumn{1}{c}{\textbf{Character 4: Sophia Lutz}} \\
    \midrule
    \texttt{\textbf{Name:} Sophia Lutz} \\
    \texttt{\textbf{Age:} 29} \\
    \texttt{\textbf{Innate:} Brave, Determined, Empathetic} \\
    \texttt{\textbf{Learned:} Sophia Lutz is a dedicated police officer who prides herself on keeping order and helping others. With the city frozen, she takes it upon herself to protect the small group of unaffected individuals and maintain some semblance of law and order.} \\
    \texttt{\textbf{Currently:} Sophia spends her days patrolling the city and ensuring the safety of the group. She has taken on the role of a leader, organizing supplies and mediating conflicts between the others.} \\
    \texttt{\textbf{Lifestyle:} Sophia's routine involves regular patrols around the group's living areas, checking on the safety of everyone, and discussing plans with Claire and Dr. Reed. She also spends time reflecting on her own role in the strange situation.} \\
    \texttt{\textbf{Living Area:} Frozen City:City Center:Police Station:Office 2} \\
    \texttt{\textbf{Daily Plan Requirement:}} \\
    \texttt{1. Patrol the city} \\
    \texttt{2. Ensure group safety} \\
    \texttt{3. Organize supplies} \\
    \texttt{4. Mediate conflicts} \\
    \\
    \midrule
    
    \multicolumn{1}{c}{\textbf{Character 5: Chris Tanaka}} \\
    \midrule
    \texttt{\textbf{Name:} Chris Tanaka} \\
    \texttt{\textbf{Age:} 34} \\
    \texttt{\textbf{Innate:} Logical, Innovative, Skeptical} \\
    \texttt{\textbf{Learned:} Chris Tanaka is a tech expert who believes the frozen time event is a result of a massive technological failure or a cyber-attack. He is determined to find a logical explanation and fix the system that he believes caused it.} \\
    \texttt{\textbf{Currently:} Chris is working with computers and electronic devices to uncover clues about the phenomenon. He frequently argues with Claire over the cause, but his tech skills are invaluable in navigating the city's systems and communications.} \\
    \texttt{\textbf{Lifestyle:} Chris spends his days hacking into systems, running diagnostics, and setting up communication networks for the group. He is usually found tinkering with devices and documenting his findings in a digital log.} \\
    \texttt{\textbf{Living Area:} Frozen City:City Center:Tech Hub:Room 5} \\
    \texttt{\textbf{Daily Plan Requirement:}} \\
    \texttt{1. Diagnose tech systems} \\
    \texttt{2. Run diagnostics} \\
    \texttt{3. Set up communications} \\
    \texttt{4. Debate theories with Claire} \\
    \\
    \midrule
    
    \multicolumn{1}{c}{\textbf{Character 6: Maya Harrison}} \\
    \midrule
    \texttt{\textbf{Name:} Maya Harrison} \\
    \texttt{\textbf{Age:} 28} \\
    \texttt{\textbf{Innate:} Passionate, Emotional, Determined} \\
    \texttt{\textbf{Learned:} Maya Harrison was in the middle of a personal crisis when time froze, arguing with her partner over a significant issue. The event has left her in emotional turmoil, and she struggles with the abrupt pause in her life.} \\
    \texttt{\textbf{Currently:} Maya is trying to make sense of her emotions and find a way to resume her life once the phenomenon ends. She helps with practical tasks but is mostly focused on finding closure to her personal issues.} \\
    \texttt{\textbf{Lifestyle:} Maya spends her days alternating between assisting Sophia with group organization and reflecting on her relationship. She journals her thoughts and keeps to herself most of the time, hoping to find a resolution.} \\
    \texttt{\textbf{Living Area:} Frozen City:Suburbs:Maple Street:House 45} \\
    \texttt{\textbf{Daily Plan Requirement:}} \\
    \texttt{1. Assist with organization} \\
    \texttt{2. Reflect on personal issues} \\
    \texttt{3. Journal thoughts} \\
    \texttt{4. Seek emotional closure} \\
    \\

    \bottomrule

    \caption{Character setup for Story 1.}
    \label{tab:story_1_character_setup}
  \end{longtable}
\end{small}

\begin{small}
  \begin{longtable}{p{\textwidth}}
    \toprule
    \multicolumn{1}{c}{\textbf{Story 1}} \\
    \midrule
    \texttt{\textbf{[Premise]}} \\
    \texttt{After a strange phenomenon causes time to freeze for everyone except for a small group of individuals, a young scientist named Claire must find a way to reverse the event before she loses her sanity.} \\\\
    \texttt{\textbf{[Setting]}} \\
    \texttt{The story takes place in a modern-day city that has suddenly fallen into an eerie state of paralysis, with the world frozen in place.} \\\\
    \texttt{\textbf{[Characters]}} \\
    \texttt{Claire Matthews: A brilliant but socially awkward physicist in her early 30s.} \\
    \texttt{Dr. Harold Reed: An older scientist and Claire's mentor.} \\
    \texttt{Tommy Harris: A troubled teenager who sees the event as a chance to escape his problems.} \\
    \texttt{Sophia Lutz: A police officer trying to maintain order in the chaos.} \\
    \texttt{Chris Tanaka: A tech expert who believes the phenomenon is a computer glitch.} \\
    \texttt{Maya Harrison: A woman who was in the middle of an argument with her partner when time froze.} \\
    \midrule

    \multicolumn{1}{c}{\textbf{Environment Setup}} \\
    \midrule
    \texttt{name: Frozen City} \\
    \texttt{description: A modern-day city that has been plunged into an eerie state of paralysis where time has frozen, leaving only a small group of individuals unaffected.} \\
    \texttt{cities:} \\
    \texttt{\ \ - name: City Center} \\
    \texttt{\ \ \ \ description: The bustling heart of the city, now eerily silent and frozen in time.} \\
    \texttt{\ \ \ \ places:} \\
    \texttt{\ \ \ \ \ \ - name: Highland Apartments} \\
    \texttt{\ \ \ \ \ \ \ \ description: A residential building where Claire Matthews resides.} \\
    \texttt{\ \ \ \ \ \ \ \ areas:} \\
    \texttt{\ \ \ \ \ \ \ \ \ \ - name: Room 704} \\
    \texttt{\ \ \ \ \ \ \ \ \ \ \ \ description: Claire Matthews' apartment, filled with scientific equipment.} \\
    \texttt{\ \ \ \ \ \ \ \ \ \ \ \ objects:} \\
    \texttt{\ \ \ \ \ \ \ \ \ \ \ \ \ \ - name: Research Desk} \\
    \texttt{\ \ \ \ \ \ \ \ \ \ \ \ \ \ \ \ description: A desk cluttered with scientific instruments and papers.} \\
    \texttt{\ \ \ \ \ \ \ \ \ \ \ \ \ \ - name: Whiteboard} \\
    \texttt{\ \ \ \ \ \ \ \ \ \ \ \ \ \ \ \ description: A whiteboard covered in equations and theories about the time freeze.} \\
    \texttt{\ \ \ \ \ \ - name: Central Library} \\
    \texttt{\ \ \ \ \ \ \ \ description: A grand library filled with books and resources, now frozen in time.} \\
    \texttt{\ \ \ \ \ \ \ \ areas:} \\
    \texttt{\ \ \ \ \ \ \ \ \ \ - name: Research Section} \\
    \texttt{\ \ \ \ \ \ \ \ \ \ \ \ description: A section filled with scientific journals and texts.} \\
    \texttt{\ \ \ \ \ \ \ \ \ \ \ \ objects:} \\
    \texttt{\ \ \ \ \ \ \ \ \ \ \ \ \ \ - name: Bookshelves} \\
    \texttt{\ \ \ \ \ \ \ \ \ \ \ \ \ \ \ \ description: Shelves containing volumes of research material.} \\
    \texttt{\ \ \ \ \ \ \ \ \ \ \ \ \ \ - name: Reading Table} \\
    \texttt{\ \ \ \ \ \ \ \ \ \ \ \ \ \ \ \ description: A table where visitors could study and read.} \\
    \texttt{\ \ \ \ \ \ - name: Tech Hub} \\
    \texttt{\ \ \ \ \ \ \ \ description: A high-tech office building where Chris Tanaka works.} \\
    \texttt{\ \ \ \ \ \ \ \ areas:} \\
    \texttt{\ \ \ \ \ \ \ \ \ \ - name: Room 5} \\
    \texttt{\ \ \ \ \ \ \ \ \ \ \ \ description: Chris Tanaka's workspace filled with computers and technical equipment.} \\
    \texttt{\ \ \ \ \ \ \ \ \ \ \ \ objects:} \\
    \texttt{\ \ \ \ \ \ \ \ \ \ \ \ \ \ - name: Server Rack} \\
    \texttt{\ \ \ \ \ \ \ \ \ \ \ \ \ \ \ \ description: A rack of servers containing important data.} \\
    \texttt{\ \ \ \ \ \ \ \ \ \ \ \ \ \ - name: Workstation} \\
    \texttt{\ \ \ \ \ \ \ \ \ \ \ \ \ \ \ \ description: A computer station set up for coding and analysis.} \\
    \texttt{\ \ \ \ \ \ - name: Police Station} \\
    \texttt{\ \ \ \ \ \ \ \ description: The main station where Sophia Lutz worked, now a base of operations.} \\
    \texttt{\ \ \ \ \ \ \ \ areas:} \\
    \texttt{\ \ \ \ \ \ \ \ \ \ - name: Office 2} \\
    \texttt{\ \ \ \ \ \ \ \ \ \ \ \ description: Sophia's office, now used for organizing group safety.} \\
    \texttt{\ \ \ \ \ \ \ \ \ \ \ \ objects:} \\
    \texttt{\ \ \ \ \ \ \ \ \ \ \ \ \ \ - name: Filing Cabinet} \\
    \texttt{\ \ \ \ \ \ \ \ \ \ \ \ \ \ \ \ description: A cabinet with case files and documents.} \\
    \texttt{\ \ \ \ \ \ \ \ \ \ \ \ \ \ - name: Radio} \\
    \texttt{\ \ \ \ \ \ \ \ \ \ \ \ \ \ \ \ description: A communication device used for coordinating with others.} \\
    \texttt{\ \ \ \ \ \ - name: Abandoned Warehouse} \\
    \texttt{\ \ \ \ \ \ \ \ description: A large, empty building now serving as Tommy Harris' hideout.} \\
    \texttt{\ \ \ \ \ \ \ \ areas:} \\
    \texttt{\ \ \ \ \ \ \ \ \ \ - name: Room 3} \\
    \texttt{\ \ \ \ \ \ \ \ \ \ \ \ description: A makeshift living space set up by Tommy.} \\
    \texttt{\ \ \ \ \ \ \ \ \ \ \ \ objects:} \\
    \texttt{\ \ \ \ \ \ \ \ \ \ \ \ \ \ - name: Sleeping Bag} \\
    \texttt{\ \ \ \ \ \ \ \ \ \ \ \ \ \ \ \ description: A sleeping bag laid out on the floor.} \\
    \texttt{\ \ \ \ \ \ \ \ \ \ \ \ \ \ - name: Backpack} \\
    \texttt{\ \ \ \ \ \ \ \ \ \ \ \ \ \ \ \ description: A backpack filled with scavenged supplies.} \\
    \texttt{\ \ \ \ \ \ - name: City Park} \\
    \texttt{\ \ \ \ \ \ \ \ description: A large, open park now frozen in a moment of stillness.} \\
    \texttt{\ \ \ \ \ \ \ \ areas:} \\
    \texttt{\ \ \ \ \ \ \ \ \ \ - name: Fountain Square} \\
    \texttt{\ \ \ \ \ \ \ \ \ \ \ \ description: A central area with a large, frozen fountain.} \\
    \texttt{\ \ \ \ \ \ \ \ \ \ \ \ objects:} \\
    \texttt{\ \ \ \ \ \ \ \ \ \ \ \ \ \ - name: Fountain} \\
    \texttt{\ \ \ \ \ \ \ \ \ \ \ \ \ \ \ \ description: A beautiful fountain with water frozen in mid-air.} \\
    \texttt{\ \ \ \ \ \ \ \ \ \ \ \ \ \ - name: Benches} \\
    \texttt{\ \ \ \ \ \ \ \ \ \ \ \ \ \ \ \ description: Wooden benches placed around the fountain.} \\
    \texttt{\ \ - name: Suburbs} \\
    \texttt{\ \ \ \ description: The quieter outskirts of the city where families reside.} \\
    \texttt{\ \ \ \ places:} \\
    \texttt{\ \ \ \ \ \ - name: Elmwood House} \\
    \texttt{\ \ \ \ \ \ \ \ description: A suburban home where Dr. Harold Reed lives.} \\
    \texttt{\ \ \ \ \ \ \ \ areas:} \\
    \texttt{\ \ \ \ \ \ \ \ \ \ - name: Unit 12} \\
    \texttt{\ \ \ \ \ \ \ \ \ \ \ \ description: Dr. Reed's home, filled with books and old research.} \\
    \texttt{\ \ \ \ \ \ \ \ \ \ \ \ objects:} \\
    \texttt{\ \ \ \ \ \ \ \ \ \ \ \ \ \ - name: Study Desk} \\
    \texttt{\ \ \ \ \ \ \ \ \ \ \ \ \ \ \ \ description: A desk with a lamp and stacks of papers.} \\
    \texttt{\ \ \ \ \ \ \ \ \ \ \ \ \ \ - name: Armchair} \\
    \texttt{\ \ \ \ \ \ \ \ \ \ \ \ \ \ \ \ description: A comfortable chair used for reading and reflection.} \\
    \texttt{\ \ \ \ \ \ - name: Maple Street} \\
    \texttt{\ \ \ \ \ \ \ \ description: A residential street where Maya Harrison lives.} \\
    \texttt{\ \ \ \ \ \ \ \ areas:} \\
    \texttt{\ \ \ \ \ \ \ \ \ \ - name: House 45} \\
    \texttt{\ \ \ \ \ \ \ \ \ \ \ \ description: Maya's home, paused in the midst of a personal argument.} \\
    \texttt{\ \ \ \ \ \ \ \ \ \ \ \ objects:} \\
    \texttt{\ \ \ \ \ \ \ \ \ \ \ \ \ \ - name: Dining Table} \\
    \texttt{\ \ \ \ \ \ \ \ \ \ \ \ \ \ \ \ description: A table with an unfinished meal.} \\
    \texttt{\ \ \ \ \ \ \ \ \ \ \ \ \ \ - name: Family Photo} \\
    \texttt{\ \ \ \ \ \ \ \ \ \ \ \ \ \ \ \ description: A photo of Maya and her partner, frozen on the mantelpiece.} \\
    \texttt{\ \ - name: Industrial District} \\
    \texttt{\ \ \ \ description: An area filled with factories and warehouses, now eerily quiet.} \\
    \texttt{\ \ \ \ places:} \\
    \texttt{\ \ \ \ \ \ - name: Old Factory} \\
    \texttt{\ \ \ \ \ \ \ \ description: An abandoned factory, now used for exploration and scavenging.} \\
    \texttt{\ \ \ \ \ \ \ \ areas:} \\
    \texttt{\ \ \ \ \ \ \ \ \ \ - name: Production Floor} \\
    \texttt{\ \ \ \ \ \ \ \ \ \ \ \ description: A large, open space with machinery frozen in time.} \\
    \texttt{\ \ \ \ \ \ \ \ \ \ \ \ objects:} \\
    \texttt{\ \ \ \ \ \ \ \ \ \ \ \ \ \ - name: Conveyor Belt} \\
    \texttt{\ \ \ \ \ \ \ \ \ \ \ \ \ \ \ \ description: A conveyor belt stopped mid-operation.} \\
    \texttt{\ \ \ \ \ \ \ \ \ \ \ \ \ \ - name: Tool Chest} \\
    \texttt{\ \ \ \ \ \ \ \ \ \ \ \ \ \ \ \ description: A chest filled with various tools and equipment.} \\
    
    \bottomrule

    \caption{Environment setup for Story 1.}
    \label{tab:story_1_environment_setup}
  \end{longtable}
\end{small}

\twocolumn

\section{Prompts}
\label{sec:prompts}

This section presents a selection of the prompts used in this paper, as shown in Table~\ref{tab:prompts_example}. These prompts play a crucial role in the sandbox initialization process, as well as in setting up the characters and environment, ensuring the system functions as intended.

For the full collection of prompts used throughout this study, which encompasses all aspects of the sandbox simulation, please refer to our project.

\onecolumn

\begin{small}
  \begin{longtable}{p{\textwidth}}
    \toprule
    \multicolumn{1}{c}{\textbf{Generate World}} \\
    \midrule

    \texttt{Variables:} \\
    \texttt{!<INPUT 0>! -- Example world.yaml} \\
    \texttt{!<INPUT 1>! -- Premise} \\
    \texttt{!<INPUT 2>! -- Setting} \\
    \texttt{!<INPUT 3>! -- Character list} \\
    \\
    \texttt{<commentblockmarker>\#\#\#</commentblockmarker>} \\
    \texttt{Below is the world.yaml file needed for use in the virtual sandbox.} \\
    \\
    \texttt{Example:} \\
    \texttt{!<INPUT 0>!} \\
    \\
    \texttt{Relevant settings of the existing story:} \\
    \texttt{Premise:} \\
    \texttt{!<INPUT 1>!} \\
    \\
    \texttt{Setting:} \\
    \texttt{!<INPUT 2>!} \\
    \\
    \texttt{Characters:} \\
    \texttt{!<INPUT 3>!} \\
    \\
    \texttt{Please generate a world.yaml file based on this setting, following the same format as the provided example, and ensure it is in YAML format:} \\
    \\
    \texttt{Return the output in the following format:} \\
    \\
    \verb|```|\texttt{yaml} \\
    \\
    \verb|```| \\
    \\
    \texttt{Please ensure that the output can be read by Python's yaml library. You only need to respond with that code block portion, without any additional content.} \\
    \\
    \texttt{Output in YAML format:} \\
    \midrule

    \multicolumn{1}{c}{\textbf{Generate Persona Scratch Information}} \\
    \midrule

    \texttt{Variables:} \\
    \texttt{!<INPUT 0>! -- Example scratch.json} \\
    \texttt{!<INPUT 1>! -- Premise} \\
    \texttt{!<INPUT 2>! -- Setting} \\
    \texttt{!<INPUT 3>! -- Character list} \\
    \texttt{!<INPUT 4>! -- World} \\
    \\
    \texttt{<commentblockmarker>\#\#\#</commentblockmarker>} \\
    \texttt{Below is the persona scratch.json file needed for use in the virtual sandbox.} \\
    \\
    \texttt{Example:} \\
    \texttt{!<INPUT 0>!} \\
    \\
    \texttt{Relevant settings of the existing story:} \\
    \texttt{Premise:} \\
    \texttt{!<INPUT 1>!} \\
    \\
    \texttt{Setting:} \\
    \texttt{!<INPUT 2>!} \\
    \\
    \texttt{Characters:} \\
    \texttt{!<INPUT 3>!} \\
    \\
    \texttt{World:} \\
    \texttt{!<INPUT 4>!} \\
    \\
    \texttt{Please note that living\_area is related to the world, starting from the root node and using colons to separate each level, with a total of four levels. For example: Frozen City:City Center:Tech Hub:Room 5} \\
    \\
    \texttt{Please generate scratch files for these characters under the given settings, formatted the same as the provided example, and in JSON format with the outer layer as a list:} \\
    \\
    \texttt{Return the output in the following format:} \\
    \\
    \verb|```|\texttt{json} \\
    \texttt{[} \\
    \texttt{\ \ \ \ \{persona 1\},} \\
    \texttt{\ \ \ \ \{persona 2\},} \\
    \texttt{\ \ \ \ ...} \\
    \texttt{]} \\
    \verb|```| \\
    \\
    \texttt{Please ensure that the output can be read by Python's JSON library. You only need to respond with that code block portion, without any additional content.} \\
    \\
    \texttt{Output in JSON format:} \\
    \midrule

    \multicolumn{1}{c}{\textbf{Generate Persona's Spatial Memory}} \\
    \midrule

    \texttt{Variables:} \\
    \texttt{!<INPUT 0>! -- Example spatial\_memory.json} \\
    \texttt{!<INPUT 1>! -- World} \\
    \\
    \texttt{<commentblockmarker>\#\#\#</commentblockmarker>} \\
    \texttt{Below is the spatial\_memory.json file needed for use in the virtual sandbox.} \\
    \\
    \texttt{Example:} \\
    \texttt{!<INPUT 0>!} \\
    \\
    \texttt{World:} \\
    \texttt{!<INPUT 1>!} \\
    \\
    \texttt{Please convert the above world.yaml file into a JSON file like the example, with the same format.} \\
    \\
    \texttt{Return the output in the following format:} \\
    \\
    \verb|```|\texttt{json} \\
    \\
    \verb|```| \\
    \\
    \texttt{Please ensure that the output can be read by Python's JSON library. You only need to respond with that code block portion, without any additional content.} \\
    \\
    \texttt{Output in JSON format:} \\

    \bottomrule

    \caption{Examples of prompts used for sandbox initialization and setup of characters and environment.}
    \label{tab:prompts_example}
  \end{longtable}
\end{small}

\twocolumn

\section{More Implementation Details}
\label{sec:more_impl_details}

The large language model used in this method is \texttt{GPT-4o mini}. For the evaluation presented in this paper, we employ the \texttt{llama3.1:8b-instruct-fp16} model.

\subsection{Reproducing Re$^3$}

The Re$^3$ method requires the specification of a story's premise, setting, and character names and descriptions. Therefore, we can directly adapt the corresponding data from the dataset to match the required format for Re$^3$. Since Re$^3$ utilizes models such as \texttt{text-davinci-002}, and to ensure fairness in the experiment, we replace these models with \texttt{GPT-4o mini}. For other hyperparameter settings, we use the default configurations provided by the method.

\subsection{Reproducing DOC-V2}

During the reproduction of DOC-V2, we observe that the prompts used in this method often result in parsing failures. Additionally, since this method does not specify the type of story, we set the story type of this method to ``narrative'' during evaluation. To ensure fairness, we replace all OpenAI engines used in this method with \texttt{GPT-4o mini}.

\subsection{Reproducing IBSEN}

During the reproduction of IBSEN, we observe that this method primarily utilizes a multi-agent virtual sandbox for generating theatrical scripts. As a result, it is necessary to first convert the premise, setting, and characters from the dataset into the specific data format required by IBSEN.

To achieve this transformation, we employ \texttt{GPT-4o mini} to automatically process the conversion. This step ensures that the data is correctly formatted and fully compatible with IBSEN's input requirements, ultimately allowing us to generate the necessary script files and other related outputs.

The final output of this method consists of log files containing dialogues between different characters. Since these logs primarily capture character interactions in a structured manner, we leverage a large language model (LLM) to further refine them into coherent and fluent stories. This post-processing step enhances readability and ensures that the generated stories flow naturally, making them suitable for evaluation and further analysis.

\subsection{Implementation Details of StoryBox}

In the multi-agent virtual sandbox simulation of this method, several hyperparameters are configurable to tailor the environment to the needs of the generation process. The simulation is initialized at the timestamp ``2024-09-01 12:00'', with each simulation step representing a time interval of one hour. We utilize \texttt{sqlite3} as the underlying database to store and manage the simulation data.

For the embedding module, we use the \texttt{jinaai/jina-embeddings-v3} model to generate embeddings, ensuring a robust representation of textual information. When integrating a large language model (LLM) into the simulation, we set the model's temperature to 0.8, with a maximum of five attempts to parse each generated output. Our observations suggest that this configuration typically allows for successful parsing within the set number of attempts. If the output cannot be successfully parsed within these attempts, the current iteration for the affected agent is skipped, but the simulation proceeds with the remaining agents.

The LLM is configured with a timeout of 60 seconds per query. The context window size is capped at 102,400 tokens, which is approximately 80\% of the model's maximum context window capacity of 128,000 tokens. This ensures that we maintain an optimal balance between context coverage and computational efficiency.

In the agent's planning module, we introduce an ``abnormal factor'' set to 0.3, meaning that there is a 30\% probability for an agent to exhibit abnormal behavior during its planning phase. This randomness is introduced to simulate unpredictable or creative decision-making, enhancing the dynamic nature of the simulation.

For the agent's execution module, the dialogue is structured to consist of two interaction cycles, resulting in a total of four conversational turns (two exchanges per agent). This configuration allows for a meaningful exchange while keeping the dialogue concise enough to maintain relevance to the ongoing story.

Finally, in the FAISS-based vector database, we set the vector dimension to 512, providing a balance between high-quality embedding representation and efficient storage and retrieval capabilities for the agent interactions.

In our experiments, GPU-dependent components, such as locally deployed embedding models and large language models, run on a single NVIDIA GTX 3090 GPU. Other models, including OpenAI's models and \texttt{DeepSeek-V3}, are accessed via API calls. When using \texttt{GPT-4o mini} for multi-agent sandbox simulations, a setup with six characters and a simulation time step of one hour requires approximately 0.5 hours of real time to simulate a full in-game day. Consequently, simulating 7 days takes around 4 hours, while a 14-day simulation requires approximately 7 hours, including both the simulation and story generation processes.

\section{Case Study}
\label{sec:case_study}

We conduct a detailed analysis of a story generated by our method, as shown in Table~\ref{tab:case_study}. In this table, we highlight four key aspects: psychological descriptions, environmental descriptions, conflicts, and resolutions. Psychological descriptions are marked in \textcolor{blue}{blue} with the prefix \textcolor{blue}{[Psychological]}, environmental descriptions in \textcolor{orange}{orange} with the prefix \textcolor{orange}{[Environment]}, conflicts in \textcolor{red}{red} with the prefix \textcolor{red}{[Conflict]}, and resolutions in \textcolor{resolu_color}{green} with the prefix \textcolor{resolu_color}{[Resolution]}. Additionally, we indicate omitted parts of the story by highlighting them in \textcolor{gray}{gray}.

The results indicate that the story contains a substantial amount of psychological and environmental descriptions, with each chapter beginning with an environmental setup. In the first chapter, conflicts are relatively sparse, primarily establishing the premise of the story. As the narrative progresses into the second and third chapters, conflicts become more frequent, marking the development phase. In the fourth chapter, both conflicts and resolutions emerge, signaling the story's climax. Finally, in the fifth chapter, conflicts and resolutions disappear, leaving only psychological and environmental descriptions, indicating the conclusion.  

Overall, the story follows a coherent structure, encompassing an introduction, development, climax, and resolution, aligning with conventional storytelling patterns.

\onecolumn

\begin{small}
  \captionsetup{width=\textwidth}
  \begin{longtable}{p{\textwidth}}
    \toprule
    \multicolumn{1}{c}{\normalsize \textbf{Frozen Echoes: Connections in a Time-Stopped World}} \\
    \midrule \\
    \multicolumn{1}{c}{\normalsize \textbf{Chapter 1: Echoes of Silence}} \\\\

    \texttt{\textcolor{blue}{[Psychological] As Claire Matthews sifted through the remnants of her mentor's cluttered office, the coldness of the environment seeped into her bones, reminding her of the chilling stillness that enveloped Frosthaven.} \textcolor{orange}{[Environment] Papers were strewn about like fallen leaves, remnants of frantic research and desperate hope.} She focused on an old, dusty notebook tucked beneath a pile of yellowed documents. Its spine cracked as she opened it, revealing a flurry of handwritten notes filled with equations and sketches of temporal mechanics. But what caught her attention was an erratic series of annotations in the margins, written with a shaky hand, suggesting a connection to the anomaly that had ensnared their city.} \\\\

    \texttt{"What is this?" she murmured, her pulse quickening. The ink was barely legible, but the words "time fracture" and "experiment 47" stood out starkly. \textcolor{blue}{[Psychological] She traced the lines with her finger, a mix of excitement and dread bubbling within her. Could this be the key to understanding what had happened?}} \\\\

    \texttt{Just then, Dr. Harold Reed walked in, his presence cutting through the silence like a beacon. "Ah, Claire. Diving into the past again, are we?" His voice carried a warmth that momentarily eased her growing anxiety.}
    \texttt{\textcolor{gray}{...[246 words]...}} \\\\

    \texttt{As she stepped out of the office, the weight of the notebook felt heavier in her hands, a tangible link to the past and a guiding light toward their uncertain future. \textcolor{orange}{[Environment] The chilling silence of Frosthaven loomed outside, but inside, Claire's mind buzzed with possibilities.} \textcolor{blue}{[Psychological] She recalled the faces of her teammates—Maya, with her emotional insights; Sophia, with her unwavering leadership; and Tommy, whose scavenged creativity could offer new perspectives.} They were bound together not just by the anomaly, but by their own stories, each seeking connection amidst the echoes of their frozen city.}
    \texttt{\textcolor{gray}{...[65 words]...}} \\\\

    \texttt{As she entered the room, she called out, "Everyone! I found something! We need to talk about 'experiment 47' and what it could mean for us!" The world outside was silent,\textcolor{blue}{[Psychological] but inside her heart, a fire was beginning to blaze.} \textcolor{orange}{[Environment] As the team gathered around the long, battered table in the dimly lit makeshift lab, \textcolor{blue}{[Psychological] Claire felt a wave of anticipation ripple through her.} The notebook lay open before them, its pages a chaotic testament to the desperation that had led to the time freeze.}}
    \texttt{\textcolor{gray}{...[67 words]...}} \\\\

    \texttt{He paused, looking each member of the team in the eye as if urging them to unearth their vulnerabilities. \textcolor{orange}{[Environment] The faint hum of the generator in the corner provided an eerie backdrop to the silence that followed, amplifying the gravity of Dr. Reed's words.} "I propose we share our experiences over the past weeks. Our personal insights might just lead us to the connections we need, both in understanding this phenomenon and in healing ourselves."} \\\\

    \texttt{\textcolor{red}{[Conflict] Maya, sitting cross-legged on her chair, looked up from her sketchbook. "But Dr. Reed, can we really afford to get sidetracked by our feelings? We need to stay focused on the science!"}}
    \texttt{\textcolor{gray}{...[33 words]...}} \\\\

    \texttt{As Maya processed this, Tommy, who had been fidgeting with a piece of scrap metal, finally spoke up. \textcolor{red}{[Conflict] "I... I can share something. The freeze didn't just stop time for me; it stopped everything. I used to love scavenging through the old tech, finding treasures in the ruins. But now... it's like everything I find is just a relic of a past I can't access. It's frustrating. I feel like I'm stuck in a still frame."}}
    \texttt{\textcolor{gray}{...[492 words]...}} \\\\
    
    \texttt{\textcolor{orange}{[Environment] The silence of Frosthaven still hung outside like a heavy fog, but within the room, a flicker of warmth had taken root, hinting at the resilience they would need to navigate the challenges that lay ahead.} As the discussion wound down, Tommy sat in silence, his eyes drifting to the corner of the makeshift lab where remnants of their scavenged materials lay haphazardly piled together. \textcolor{orange}{[Environment] The metallic glint of circuit boards, the soft sheen of unbroken glass, and scraps of colorful plastic seemed to whisper to him, urging him to create something meaningful out of the stillness surrounding them.}}
    \texttt{\textcolor{gray}{...[508 words]...}} \\\\

    \texttt{\textcolor{blue}{[Psychological] Claire felt a warmth spread through her chest as she watched her teammates collaborate, their energies intertwining like threads in a fabric.} "This is what we need, isn't it? To remember that even in the face of this freezing silence, we're still here, still human."} \\\\

    \texttt{As they stood amidst their creation, the once-frozen city outside felt a bit warmer, the silence a little less oppressive. In that moment, they understood that art—their shared expression of grief and hope—was a powerful tool, one that could bridge the chasm between their isolation and the humanity they longed to reconnect with.} \\\\

    \texttt{With the installation complete and plans for a community gathering taking shape, \textcolor{blue}{[Psychological] Claire felt an invigorating breeze of hope swell within her.} The journey towards understanding the anomaly had begun, and with it, their paths intertwined ever more deeply, ready to face whatever challenges lay ahead together.} \\\\

    \midrule \\
    \multicolumn{1}{c}{\normalsize \textbf{Chapter 2: Fractured Connections}} \\\\
    
    \texttt{\textcolor{orange}{[Environment] The afternoon sun cast a pale light over the workshop where the team congregated, illuminating the remnants of their collaborative art installation.} Claire, with sketches scattered around her, stood by the large easel, her brow furrowed in concentration as she reviewed the group's progress. The atmosphere was thick with anticipation, but tensions simmered beneath the surface, ready to boil over.} \\\\

    \texttt{\textcolor{red}{[Conflict] Chris, immersed in his laptop, suddenly slammed the lid shut, breaking the silence. "We can't keep talking about feelings, Claire! We need data, not emotional fluff! This is about survival, not therapy!" His voice rose, echoing off the cold concrete walls, causing heads to turn.}} \\\\

    \texttt{\textcolor{blue}{[Psychological] Claire's heart raced.} \textcolor{red}{[Conflict] "How can you say that? Emotions are part of our experience! We can't just shove them aside as if they don't matter! If we don't understand the human element, how do we hope to solve the anomaly?"}}
    \texttt{\textcolor{gray}{...[40 words]...}} \\\\

    \texttt{\textcolor{red}{[Conflict] Chris scoffed, shaking his head. "You're both missing the bigger picture! The anomaly is a scientific problem, and we need to treat it as such. Statistics and models, that's what we need! Not a group therapy session!"}} \\\\

    \texttt{\textcolor{orange}{[Environment] A heavy silence followed his outburst, filled only by the distant hum of the city's systems, eerily frozen in time.} \textcolor{blue}{[Psychological] Claire felt a weight settle in her chest, her frustration mixing with the isolation she had been struggling with since the freeze began.} She took a deep breath, grounding herself. \textcolor{red}{[Conflict] "I'm not asking for therapy, Chris. I'm asking for understanding. All of us are feeling the strain of this anomaly. If we don't address the emotional fallout, we risk losing more than just our city—we risk losing ourselves."}}
    \texttt{\textcolor{gray}{...[39 words]...}} \\\\

    \texttt{\textcolor{red}{[Conflict] Chris threw his hands up, cutting him off. "That's ridiculous, Tommy! This isn't a support group! We're not here to share our feelings; we're here to fix a problem!" His anger was palpable, and the room tensed at the ferocity of his words.}}
    \texttt{\textcolor{gray}{...[54 words]...}} \\\\

    \texttt{Chris clenched his jaw, visibly wrestling with his emotions as he paced the room, his agitation bouncing off the walls. \textcolor{blue}{[Psychological] The others watched, feeling the weight of the moment.} Finally, he turned to Claire, his tone softening slightly. "I just... I can't afford to lose focus. We're running out of time."} \\\\

    \texttt{\textcolor{blue}{[Psychological] Claire's heart softened at his vulnerability.} "None of us can afford to lose focus, Chris. But we're not in this alone, and we shouldn't have to carry it all on our shoulders. What if we tried combining both approaches? We could analyze the data, but also create a space for sharing. We could document our emotional states alongside our findings. It might help us see patterns we wouldn't have considered otherwise."}
    \texttt{\textcolor{gray}{...[81 words]...}} \\\\

    \texttt{As the discussion shifted towards finding a framework that honored both their emotional and scientific needs, Claire felt a warmth spread through the group. \textcolor{blue}{[Psychological] They were angry, frustrated, and scared, but they were also moving toward a shared understanding, forging a deeper connection amid the chaos.}}
    \texttt{\textcolor{gray}{...[903 words]...}} \\\\

    \texttt{\textcolor{orange}{[Environment] As the sun dipped below the horizon, casting a soft golden glow over Frosthaven, Sophia stood in the community center, surveying the room with a mixture of excitement and apprehension. The large, open space was adorned with strings of fairy lights, their warmth contrasting with the cold, sterile environment of their scientific workspace. Canvas draped across tables awaited contributions, and Maya's masterpiece, "Isolation's Embrace," was prominently displayed at the front, its swirling colors capturing the emotional struggles faced by the community.}}
    \texttt{\textcolor{gray}{...[305 words]...}} \\\\

    \texttt{\textcolor{red}{[Conflict] When it was Chris's turn to speak, he reluctantly approached the makeshift stage, a shadow of frustration crossing his features. "I appreciate the intent behind this gathering, but we have to remember that feelings don't solve the problem. We need actionable data—statistics, research, something concrete to work with!"}}
    \texttt{\textcolor{gray}{...[405 words]...}} \\\\

    \texttt{\textcolor{red}{[Conflict] Chris sighed, his shoulders slumping. "I just don't want to lose focus. If we get too caught up in emotions, we might miss our chance to save the city."}}
    \texttt{\textcolor{gray}{...[35 words]...}} \\\\

    \texttt{As they exited the community center, \textcolor{blue}{[Psychological] Claire felt the weight of the evening's revelations.} While the gathering had illuminated hidden tensions, it had also sparked connections that could lead them toward healing. They were still a fractured group, but together, they could navigate the complexities of their emotions and unite in their quest to reclaim Frosthaven. The road ahead would be challenging, but they were beginning to understand that it was possible to blend their individual desires with a greater purpose—one that could ultimately drive them toward collective resilience.} \\\\

    \midrule \\
    \multicolumn{1}{c}{\normalsize \textbf{Chapter 3: The Heart of Inquiry}} \\\\
    
    \texttt{\textcolor{orange}{[Environment] The air in the makeshift lab crackled with tension as Chris stared at his computer screen, the bright glow illuminating his furrowed brow. A cacophony of beeping alarms punctuated the silence that had settled uneasily over the group, a sound that mirrored the rising pulse of anxiety coursing through everyone present.} Claire, Maya, Sophia, and Tommy stood gathered around him, their expressions a blend of concern and disbelief as the reality of Chris's actions unfolded before them.} \\\\

    \texttt{\textcolor{red}{[Conflict] "What have you done?" Claire's voice trembled slightly, a mixture of fear and frustration. She stepped closer, her mind racing with the implications of the power surge.}} \\\\

    \texttt{\textcolor{red}{[Conflict] Chris, his hands still hovering over the keyboard, shot her a defiant glance. "I was just trying to access the central grid! If we can understand the frozen energy signatures, maybe we can find a way to reverse this freeze. We can save Frosthaven!"}}
    \texttt{\textcolor{gray}{...[49 words]...}} \\\\

    \texttt{\textcolor{orange}{[Environment] The monitors flickered, reflecting a kaleidoscope of error messages and warnings, each one a testament to the risk they were all contemplating.} \textcolor{red}{[Conflict] Tommy leaned over Chris's shoulder, his eyes darting between the screen and the faces of his friends. "We should think this through. What if we lose everything?"} He was usually the one advocating for the thrill of experimentation, but this was different. The stakes felt heavier now.} \\\\

    \texttt{Sophia, who had been quietly observing, stepped forward. Her voice was steady, an anchor amidst the storm. \textcolor{red}{[Conflict] "We need to decide if this is the right path. We're already facing the consequences of our actions—what's one more gamble on top of all this?"} Her gaze swept across the group, searching for consensus. "Maybe there's another way. Let's take a moment to regroup and consider all our options."} \\\\

    \texttt{\textcolor{blue}{[Psychological] Claire nodded, her heart pounding as she recalled their earlier discussions about balancing emotional clarity with scientific inquiry.} "I agree with Sophia. This isn't just about reversing the anomaly. It's about us too. We need to find a solution that doesn't put any of you at risk."}

    \texttt{\textcolor{red}{[Conflict] Chris's face hardened at their resistance, yet a flicker of uncertainty crossed his features. "But we don't have time to sit around! Every moment we waste, more lives are impacted. People are trapped in this stasis!"} His voice rose, desperation lacing his words, revealing the turmoil beneath his bravado.}
    \texttt{\textcolor{gray}{...[50 words]...}} \\\\

    \texttt{Claire watched Chris as his expression shifted, the wall of bravado beginning to crack. The tension in the room thickened, silence stretching out like an elastic band ready to snap. \textcolor{blue}{[Psychological] Finally, Chris slumped back in his chair, his anger dissipating into a weary resignation.} "I just thought... maybe this was the way to prove I'm not just a tech geek who hides behind a screen. I wanted to be part of something bigger."}
    \texttt{\textcolor{gray}{...[30 words]...}} \\\\

    \texttt{The group sat in contemplative silence, the weight of their shared isolation settling among them. \textcolor{blue}{[Psychological] Claire inhaled deeply, her resolve strengthening.} "Let's take a step back. We can analyze the data from your hack without triggering further consequences. We can create a simulation. Test it. Navigate the risks together."} \\\\

    \texttt{As they discussed their next steps, the room began to hum with a renewed sense of collaboration. They drew on each other's strengths, merging Claire's scientific expertise, Maya's creative intuition, and Sophia's leadership skills. \textcolor{blue}{[Psychological] Chris, though still filled with frustration, felt the warmth of camaraderie seeping back into the frigid edges of his heart.}} \\\\

    \texttt{With the group united, they set to work, sketching out a plan that combined their talents with caution. \textcolor{orange}{[Environment] As they delved deeper into the mystery of the freeze, the flickering monitors transformed from harbingers of chaos to beacons of hope,} reflecting the strength found in their shared determination. \textcolor{red}{[Conflict] This moment marked a turning point—an understanding that every challenge they faced could either drive them apart or forge unbreakable bonds, and they chose the latter.}}
    \texttt{\textcolor{gray}{...[322 words]...}} \\\\

    \texttt{\textcolor{orange}{[Environment] As she finished reading, silence enveloped the room, but it wasn't the same tense silence as before. Instead, it felt warm, inviting a space for reflection.} Chris shifted in his seat, his face softening as he took in her words. "I never thought of it that way, Maya. I've been so focused on the data, on proving myself, that I forgot we're all human here, feeling the same fears. It's easy to hide behind numbers and screens, but I miss feeling connected."}
    \texttt{\textcolor{gray}{...[54 words]...}} \\\\

    \texttt{\textcolor{blue}{[Psychological] Claire felt a wave of relief wash over her as she realized the significance of Maya's honesty.} "This is what I've been trying to express. Our emotional journeys are just as important as our scientific ones. If we want to move forward, we need to create a safe space where we can be open with each other. We're not just colleagues; we're a team that needs to lean on one another."}
    \texttt{\textcolor{gray}{...[292 words]...}} \\\\

    \texttt{\textcolor{orange}{[Environment] As the night deepened, the flickering lights of the makeshift lab cast long shadows, elongating the figures of Claire Matthews and Dr. Harold Reed as they huddled over a table strewn with half-filled coffee cups and scattered notes. Outside, the frozen city of Frosthaven lay still, a haunting reminder of the urgency that fueled their late-night discourse.} The tension from earlier in the evening had dissipated, replaced by a palpable sense of purpose.}
    \texttt{\textcolor{gray}{...[714 words]...}} \\\\

    \texttt{\textcolor{orange}{[Environment] When they finally stepped away from the table, the energy in the room felt charged, like the calm before a storm of creativity and collaboration.} Claire turned to Dr. Reed, her eyes glinting with determination. "Let's bring this to the team. I believe together, we can thaw the emotional freeze and reignite the spirit of Frosthaven."} \\\\

    \texttt{As they prepared to share their newfound idea at the next meeting, \textcolor{blue}{[Psychological] Claire felt a thrill of anticipation and the kindling of unity among her friends.} They were not just researchers anymore; they were storytellers, artists, and empathizers—ready to weave the rich tapestry of their experiences into the heart of their mission. This chapter in their journey marked not just a scientific inquiry, but the dawning of a renewed sense of community, where every voice mattered, and every story echoed with purpose.} \\\\

    \midrule \\
    \multicolumn{1}{c}{\normalsize \textbf{Chapter 4: Frozen Reflections}} \\\\
    
    \texttt{As the group settled into the warmth of their makeshift retreat, the ambiance was both somber and reflective. \textcolor{orange}{[Environment] The room, dimly lit by flickering candles, held remnants of their earlier discussions—a few sketches from Maya, hastily written notes from Tommy, and a whiteboard filled with diagrams and equations. Outside, the silent city of Frosthaven loomed, a ghostly reminder of their shared plight.}} \\\\

    \texttt{Claire shifted uneasily in her seat, glancing at the others. Her heart raced as she sensed the stories hovering just beneath the surface, waiting to be shared. "You know," she began, her voice soft yet steady, "we all ended up here for a reason. But sometimes I wonder—what choices brought us to this frozen moment?"}
    \texttt{\textcolor{gray}{...[1474 words]...}} \\\\

    \texttt{\textcolor{red}{[Conflict] "We can't just keep talking about feelings and art! We need a solid plan if we want any chance of reversing the anomaly. The science is what matters right now, not these sentimental projects!" His words cut through the room, sharp and jarring against the hopeful energy.}} \\\\

    \texttt{\textcolor{red}{[Conflict] Maya, taken aback, responded defensively, "Chris, this is part of the plan! If we don't connect emotionally as a team, how can we expect to tackle the scientific challenges? You're dismissing what we're trying to build!"}} \\\\

    \texttt{\textcolor{red}{[Conflict] Tommy, feeling the tension crackle in the air, hesitated between the two perspectives. "But we've seen that our emotions influence our work. If we don't acknowledge that, we might just be running in circles. We need this... connection. It's what makes us human."}} \\\\

    \texttt{\textcolor{red}{[Conflict] Chris shook his head vehemently. "Humanity is what got us into this mess! We need to focus on the cold, hard facts if we want to get out of it!"}} \\\\

    \texttt{\textcolor{blue}{[Psychological] The argument escalated, voices rising as frustration spilled over. Claire's heart raced; she could feel the rift growing within the group, the very thing they had fought to overcome beginning to crack. In this moment of chaos, she felt a familiar pang of loneliness, as if the very silence outside had crept into their hearts and taken hold.}} \\\\

    \texttt{Then, just as it seemed the discussion might devolve into a shouting match, Sophia, who had been quietly observing, interjected with a calm yet authoritative presence.} \\\\

    \texttt{\textcolor{red}{[Conflict] "Enough!" she said, her voice steady and clear, cutting through the tension like a knife. All eyes turned to her, the room falling silent. "This isn't about one approach being more valid than another. We are all here because we want to overcome the anomaly together, and we need each other to do that. What we're creating is not just art; it's a representation of our collective journey. Each of us is valid, and so are our emotions and our scientific inquiries."}}
    \texttt{\textcolor{gray}{...[98 words]...}} \\\\

    \texttt{\textcolor{resolu_color}{[Resolution] Sophia stepped closer, her gaze steady. "And I understand that fear. It's real for all of us. But remember, fear can't be the only driver of our actions. We need to harness that fear and turn it into something constructive. Let's combine our strengths. Can we agree to collaborate on both the art and the science? This isn't about choosing one over the other; it's about integrating everything we have at our disposal."}} \\\\

    \texttt{\textcolor{resolu_color}{[Resolution] The honesty in her words began to soften the resolve in Chris's expression. "So, you're saying we can use the art as a means to inform our research?"}} \\\\

    \texttt{\textcolor{resolu_color}{[Resolution] "Exactly!" Maya chimed in, her voice rising with newfound enthusiasm. "If we visualize our findings, it could help make the data more relatable—accessible—especially to the community. We can overlay the emotional narratives we've discussed with the scientific data we gather. It'll be a full-bodied representation of our journey."}} \\\\

    \texttt{With Sophia's leadership reestablished, the group began to nod in agreement. \textcolor{blue}{[Psychological] Claire felt the tension ease, the flickering candlelight dancing around them now feeling warm instead of ominous.}}
    \texttt{\textcolor{gray}{...[45 words]...}} \\\\

    \texttt{\textcolor{resolu_color}{[Resolution] As they shifted their focus back to their collaborative project, Claire's heart swelled with gratitude for Sophia's ability to unify them. They moved forward, sketching ideas and generating plans that would seamlessly weave together their emotional and scientific pursuits. The earlier argument faded into the background, replaced by a renewed sense of purpose. They were a team once more, ready to tackle the challenges that lay ahead together.}} \\\\

    \texttt{\textcolor{blue}{[Psychological] In that moment, Claire felt the weight of isolation lift, realizing that they weren't just fighting against the freeze; they were fighting for each other. United, they would explore every avenue to thaw not only the city but the connections that had brought them together in this frozen time.}} \\\\

    \midrule \\
    \multicolumn{1}{c}{\normalsize \textbf{Chapter 5: Thawing Resilience}} \\\\

    \texttt{\textcolor{orange}{[Environment] The atmosphere in the makeshift lab was thick with anticipation as Claire stood before her assembled team, a faint hum of energy pulsing through the air. The walls, adorned with sketches and diagrams born from their collective efforts, felt alive with possibility.} \textcolor{blue}{[Psychological] Each of them had poured their hearts into this moment, and now, standing at the precipice of discovery, Claire could feel the weight of their shared hopes resting on her shoulders.}} \\\\

    \texttt{"Alright, everyone, let's gather around," Claire called out, her voice steady despite the flutter of anxiety in her chest. A mix of determination and vulnerability glimmered in her eyes as she gestured to the whiteboard littered with equations and emotional notes. "I've been thinking about our last discussion—how we need to merge the emotional with the scientific. It's time we made that a reality."}
    \texttt{\textcolor{gray}{...[1697 words]...}} \\\\

    \texttt{"Okay, let's do this," Chris announced, his voice steadying the group as he clicked the final command. \textcolor{orange}{[Environment] The monitors blinked to life, displaying a flurry of colors and patterns that mirrored their emotional states.}} \\\\

    \texttt{\textcolor{orange}{[Environment] At first, the screen showed a chaotic mix of reds and blues—confusion and fear swirling together. But as the simulation progressed, the colors began to shift, slowly transforming into vibrant hues of green and gold.} \textcolor{blue}{[Psychological] Claire's heart raced as she realized what was happening: the emotional resonance they had captured was beginning to have a tangible effect on the anomaly.}}
    \texttt{\textcolor{gray}{...[272 words]...}} \\\\

    \texttt{\textcolor{orange}{[Environment] As the team took a collective breath, the thawing in the city continued, their newfound awareness grounding them even as they reveled in the joy of discovery. With each passing moment, the icy grip on Frosthaven began to wane, revealing glimpses of life that had been frozen in time.}} \\\\

    \texttt{Claire turned to her team, a fierce determination igniting in her. "Let's document everything. Our findings—the emotional model, the data, and the city's response. We need to understand how our connections are shaping this thaw. It's not just about stopping the freeze. It's about healing ourselves and our community."}
    \texttt{\textcolor{gray}{...[92 words]...}} \\\\

    \texttt{\textcolor{blue}{[Psychological] As the visuals danced on the screen, Claire caught sight of the cityscape—once lifeless, it now thrummed with the potential of awakening. She felt a surge of emotion, a bittersweet reminder of how far they had come. But she also recognized that this was just the beginning.}} \\\\

    \texttt{"Let's keep pushing," she urged, her voice rising above the excitement. "As we integrate more data, we'll invite the community to share their experiences. This isn't just our story; it's a collective one. Together, we can truly understand and harness the emotional energy that's unlocking the city."} \\\\

    \texttt{The team rallied around her words, their spirits invigorated by the vision of what lay ahead. With the city beginning to thaw, they were not only on the precipice of discovery but were also embedded in a transformative journey that would redefine their lives and the essence of Frosthaven itself.}

    \\
    \bottomrule

    \caption{Color-coded annotations in the generated story. Psychological descriptions are highlighted in \textcolor{blue}{blue} (\textcolor{blue}{[Psychological]}), environmental descriptions in \textcolor{orange}{orange} (\textcolor{orange}{[Environment]}), conflicts in \textcolor{red}{red} (\textcolor{red}{[Conflict]}), and resolutions in \textcolor{resolu_color}{green} (\textcolor{resolu_color}{[Resolution]}). Omitted parts of the story are indicated in \textcolor{gray}{gray}.}
    \label{tab:case_study}
  \end{longtable}
\end{small}

\end{document}